\documentclass[10pt,journal,compsoc]{IEEEtran}

\ifCLASSOPTIONcompsoc
  \usepackage[nocompress]{cite}
\else
  \usepackage{cite}
\fi

\let\shortcite=\cite
\usepackage{graphicx}
\usepackage{amsmath}
\usepackage{amssymb}
\usepackage{color}
\usepackage{comment}
\usepackage{booktabs}
\usepackage{subcaption}
\usepackage{tabularx}
\usepackage{stmaryrd}
\usepackage{mathrsfs}
\usepackage{url}

\DeclareMathOperator*{\ArgMin}{arg\,min}

\DeclareMathOperator*{\ArgMax}{arg\,max}

\definecolor{ColorGray}{RGB}{96, 96, 96}
\definecolor{ColorDarkGreen}{RGB}{19, 179, 50}
\definecolor{ColorLightBlue}{RGB}{18, 137, 255}
\definecolor{ColorOrange}{RGB}{252, 111, 3}

\newcommand{\rev}[1]{#1}

\newcommand{\Figure}[1]{Fig.~#1}
\newcommand{\Table}[1]{Table~#1}
\newcommand{\Equation}[1]{Eq.~#1}
\newcommand{\Section}[1]{Sec.~#1} 
\newcommand{\Etal}[1]{#1 et al.}
\newcommand{\Wrt}{w.r.t.}
\newcommand{\Ie}{i.e.}
\newcommand{\Eg}{e.g.}

\def\OurMethodName{WSDesc}

\def\Descriptor{\mathbf{f}}
\def\DescriptorDimension{n}

\def\PointCloudP{\mathcal{P}}
\def\PointCloudPMat{\mathbf{P}}
\def\PointCloudQ{\mathcal{Q}}
\def\PointCloudQMat{\mathbf{Q}}
\def\PointP{\mathbf{p}}
\def\PointQ{\mathbf{q}}
\def\Neighborhood{\mathcal{S}}
\def\NeighborhoodRadius{r^{\text{\tiny{LRF}}}}
\def\LRFTransform{\mathbf{T}^{\text{\tiny{LRF}}}}

\def\VoxelGrid{\mathcal{V}}
\def\VoxelGridResolution{h}
\def\Voxel{v}

\def\VoxelGridScale{s}
\def\VoxelRadius{r}
\def\VoxelCenter{\mathbf{o}}
\def\VoxelPointProbability{p}
\def\VoxelPointDistance{d}

\def\ConvNet{\mathscr{G}}
\def\ConvNetParams{\Phi}

\def\PointCloudTransform{\mathscr{T}}

\def\PointCloudRotation{\mathbf{R}}
\def\PointCloudTranslation{\mathbf{t}}
\def\PointCloudRotationInv{\PointCloudRotation'}
\def\PointCloudTranslationInv{\PointCloudTranslation'}

\def\CorrespondenceSet{\mathbf{C}}
\def\Correspondence{c}
\def\CompatibilityMatrix{\mathbf{M}}
\def\CorrespondenceLengthDifference{d}
\def\CorrespondenceWeightMat{\mathbf{W}}
\def\CorrespondenceWeight{\mathbf{w}}
\def\CorrespondenceWeightEntry{w}

\def\Loss{\mathcal{L}}
\def\LossRegistration{\Loss_{pcr}}
\def\LossOrthogonality{\Loss_{o}}
\def\LossOrthogonalityWeight{\lambda_{o}}

\def\LossCycle{\Loss_{c}}
\def\LossCycleWeight{\lambda_{c}}

\def\IdentityMatrix{\mathbf{I}}

\def\PointCloudPairSet{\mathbf{\Lambda}}
\def\DescriptorExtraction{\mathscr{D}}
\def\PutativeCorrespondenceSet{\mathbf{\Omega}}
\def\InlierThresh{\tau_{1}}
\def\InlierRatioThresh{\tau_{2}}

\def\GTCorrespondenceSet{\mathbf{\Omega}^{*}}

\begin{document}

\title{\OurMethodName{}: Weakly Supervised 3D Local Descriptor Learning for Point Cloud Registration}

\author{Lei~Li,
        Hongbo~Fu,
        and~Maks~Ovsjanikov
\IEEEcompsocitemizethanks{\IEEEcompsocthanksitem L. Li and M. Ovsjanikov are with 
LIX, \'{E}cole Polytechnique, IP Paris, France. 
E-mail: \{lli,maks\}@lix.polytechnique.fr\protect\\
\IEEEcompsocthanksitem H. Fu is with the School of Creative Media, 
City University of Hong Kong, Hong Kong.
E-mail: hongbofu@cityu.edu.hk}
\thanks{(Corresponding author: Hongbo Fu.)}
}

\markboth{}%
{}

\IEEEtitleabstractindextext{%
\begin{abstract}
  In this work, we present a novel method called \OurMethodName{} to learn 3D local descriptors in a weakly supervised manner for robust point cloud registration. Our work builds upon recent 3D CNN-based descriptor extractors, which leverage a voxel-based representation to parameterize local geometry of 3D points. Instead of using a predefined fixed-size local support in voxelization, we propose to \emph{learn} the optimal support in a data-driven manner. To this end, we design a novel differentiable voxelization layer that can back-propagate the gradient to the support size optimization. To train the extracted descriptors, we propose a novel registration loss based on the deviation from rigidity of 3D transformations, and the loss is weakly supervised by the prior knowledge that the input point clouds have partial overlap, without requiring ground-truth alignment information. Through extensive experiments, we show that our learned descriptors yield superior performance on existing geometric registration benchmarks.
\end{abstract}

\begin{IEEEkeywords}
Point cloud, 3D local descriptor, geometric registration, differentiable voxelization, 3D CNN, weak supervision.
\end{IEEEkeywords}}

\maketitle

\IEEEdisplaynontitleabstractindextext

\IEEEpeerreviewmaketitle

\IEEEraisesectionheading{\section{Introduction}\label{sec:introduction}}

\begin{figure}[t]
	\centering
	\includegraphics[width=\linewidth]{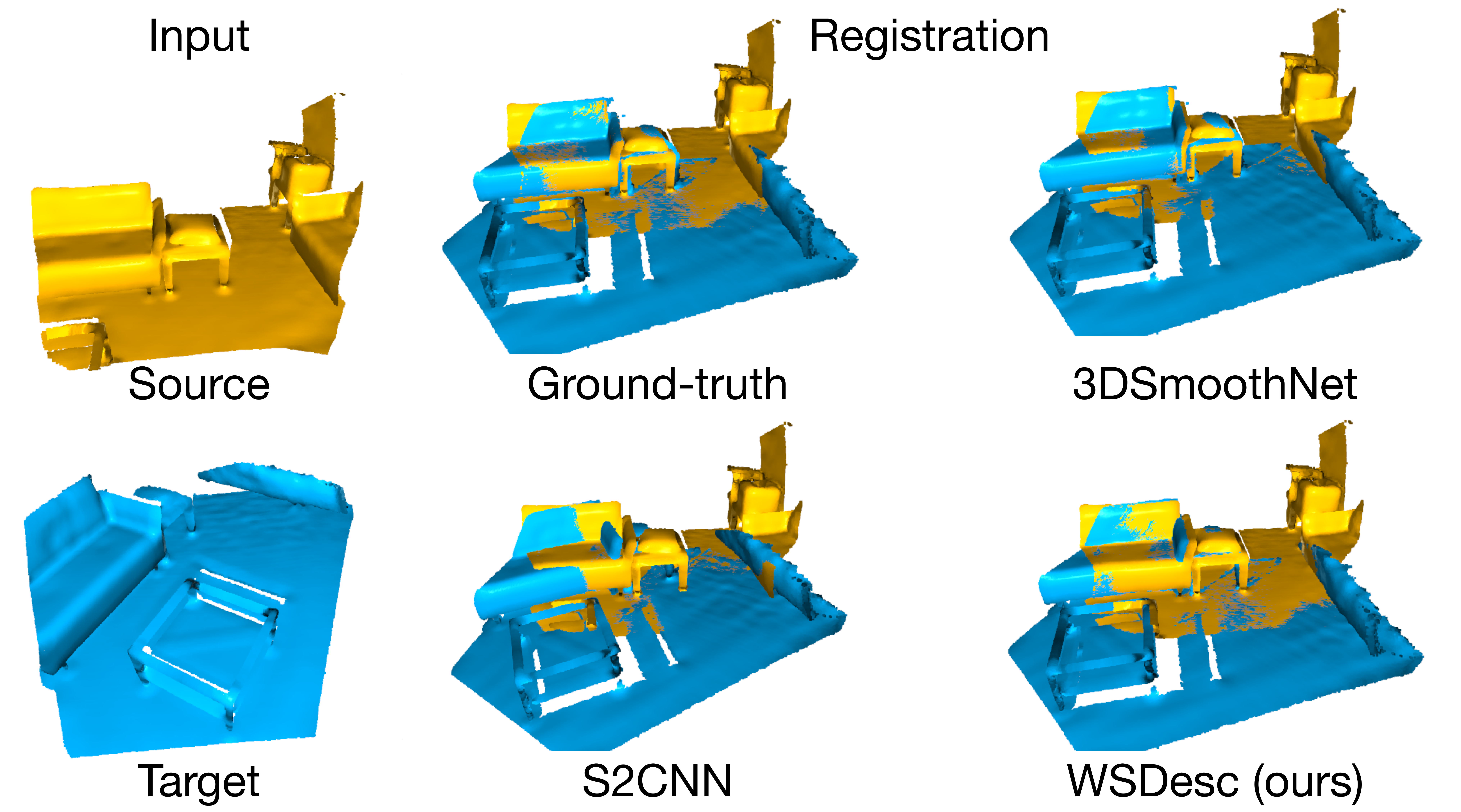}
	\caption{Registration of two point cloud fragments (with a partial overlap ratio of 51\%) using learned descriptors. 3DSmoothNet~\cite{Gojcic_2019_CVPR} is trained \emph{with} ground-truth alignment information, while S2CNN~\cite{Spezialetti_2019_ICCV} and our \OurMethodName{} do not require such information for training.}
	\label{fig:teaser}
\end{figure}

\IEEEPARstart{E}{ncoding} 3D local geometry into descriptors has been an essential ingredient in many computer graphics and vision problems,
such as recognition~\cite{Guo2013}, retrieval~\cite{Bronstein:2011:ShapeGoogle}, segmentation~\cite{Evangelos:2010:L3DMSL}, registration~\cite{mian2006novel,tam2012registration}, etc.
In this work, we are interested in developing 3D local descriptors for robust point cloud registration (\Figure{\ref{fig:teaser}}).
Matching 3D geometry of scans of real-world scenes is a challenging task due to the presence of noise and partiality in the input data.
To address such issues, learning-based descriptors have received significant attention in recent years, demonstrating superior performance over hand-crafted ones~\cite{Guo:2016:CPELFD}.

To capture local geometric structures, an important step for learning-based descriptors~\cite{Zeng_2017_CVPR,guerrero2018pcpnet,Gojcic_2019_CVPR,ao2020spinnet} is to extract local neighborhood support of 3D points with some predefined size. 
The local neighborhoods have a variety of representations, such as the point-based~\cite{guerrero2018pcpnet}, voxel grids~\cite{Gojcic_2019_CVPR}, histograms~\cite{Khoury_2017_ICCV}, or point pair features~\cite{Deng_2018_CVPR}, which are amenable to feature learning with networks.
In all of these approaches, the support size is crucial in determining the amount of local geometry information captured in the learned descriptors.
Normally, this parameter is set empirically and is not involved in network training, which may keep the network from extracting more informative descriptors.
Moreover, simply using a large support may lead to descriptors that are too global and thus specific to a shape or a scene, whereas using a small support may lead to loss of robustness and informativeness, as discussed in~\cite{Guo:2016:CPELFD,Li_2020_CVPR}.
Thus, finding an optimal support is not an effortless task.

To train local descriptors, prior works have leveraged contrastive learning, such as with the triplet~\cite{Gojcic_2019_CVPR} or the N-tuple loss~\cite{Deng_2018_CVPR}, to optimize the descriptor similarity.
Researchers have also investigated end-to-end registration-based training by applying the extracted descriptors to in-network alignment estimation for pairwise point clouds~\cite{Wang:PRNet:2019,Wang_2019_ICCV}.
In the aforementioned works~\cite{Gojcic_2019_CVPR,Deng_2018_CVPR,Wang:PRNet:2019,Wang_2019_ICCV}, the training typically requires ground-truth alignment information for supervision.
While the ground-truth can potentially be derived from existing 3D reconstruction pipelines~\cite{Choi_2015_CVPR}, it can be prone to errors, and typically requires careful manual verification.
Recently there is a growing body of literature advocating descriptor learning without supervision~\cite{Deng_2018_ECCV,Zhao_2019_CVPR,Spezialetti_2019_ICCV,Yang_2021_CVPR,Feng_2021_CVPR}, which can sidestep the issues that arise in ground-truth labeling and potentially benefit from broader untapped 3D data.

In this paper, we propose a weakly supervised 3D local descriptor learning method (\emph{\OurMethodName{}} for short),  
endowed with differentiable voxelization for a learnable local support and a novel registration loss for training without ground-truth alignment information.

Specifically, given a pair of 3D point clouds as input, to extract point descriptors, we build upon 3DSmoothNet~\cite{Gojcic_2019_CVPR}, a 3D CNN-based architecture using a voxel-based representation for 3D local geometry.
That work adopts a predefined local support for input voxelization.
In contrast, we enable the network to learn the support size in a data-driven manner.
In order to back-propagate the gradient to the support size optimization, we propose a differentiable voxelization layer to bridge the gap between the point clouds and their local voxel-based representations.

Next, to train the descriptor extraction network, we introduce a powerful registration loss based on deviation from rigidity of the alignment of the point cloud pair.
Prior works~\cite{Wang:PRNet:2019,Wang_2019_ICCV} normally formulate a registration-based loss by evaluating the difference between the ground-truth and a 3D transformation computed by matching the descriptors of the input point clouds and applying the differentiable singular value decomposition (SVD).
The use of the SVD ensures the rigidity of the estimated 3D transformation.
Differently, we propose to relax the alignment to be an \emph{affine} transformation, which can be solved in a linear system of equations without additional constraints.
Inspired by recent non-rigid correspondence techniques \cite{roufosse2019unsupervised}, in our proposed registration loss, we enforce the rigidity of the computed affine transformation by promoting its structural properties such as orthogonality and cycle consistency.

Our registration loss is \emph{weakly supervised} in the sense that we only expect the input point clouds to have partial overlap, which is to ensure the \emph{presence} of some underlying rigid alignment between the point clouds.
Notably, our registration loss does not require the knowledge of ground-truth alignment information, compared to the above SVD-based works, thus giving rise to a simpler training procedure.

Our main contributions are summarized as follows:
(1) We propose a differentiable voxelization layer, enabling in-network conversion from point clouds to a voxel-based representation and allowing data-driven local support optimization;
(2) We propose a weakly supervised registration loss based on the deviation from rigidity of 3D transformations between point clouds, effectively guiding the descriptor similarity optimization;
(3) Our method shows superior performance on existing geometric registration benchmarks.
Our code will be made publicly available\footnote{\url{https://github.com/craigleili/WSDesc}}.

\begin{figure*}[ht]
	\centering
	\includegraphics[width=\linewidth]{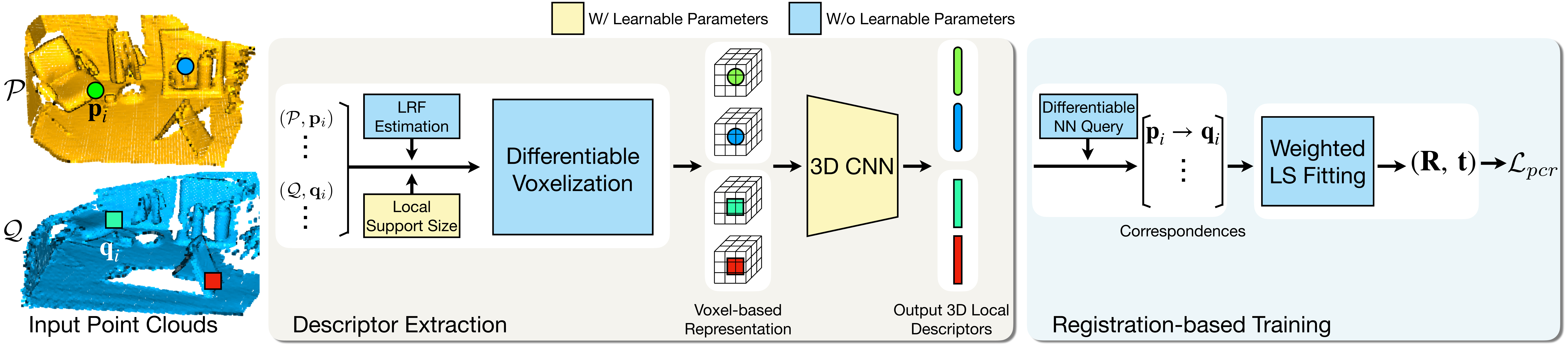}
	\caption{Overview of our fully differentiable learning pipeline  \OurMethodName. In the descriptor extraction stage, the local geometry of a point $\PointP_i$ is converted to a voxel-based representation for the 3D CNN-based descriptor extractor. Each voxel grid is transformed by a local reference frame (LRF) for rotation invariance and fitted to a learned local support size. The online point-to-voxel conversion is enabled by the differentiable voxelization layer. In the registration-based training stage, the 3D local descriptors are used for building putative correspondences between point clouds $\PointCloudP$ and $\PointCloudQ$ and estimating an affine transformation $(\PointCloudRotation, \PointCloudTranslation)$ by weighted least-squares fitting for our registration loss $\LossRegistration$.}
	\label{fig:pipeline}
\end{figure*}

\section{Related Work}
\label{sec:related-work}

In this section, we briefly discuss relevant works on both hand-crafted and learned 3D local descriptors as well as learning geometric registration.

\textbf{Hand-crafted 3D Local Descriptors.}
A considerable amount of literature has investigated hand-crafted 3D local descriptors.
A histogram-based representation is widely used to parameterize local geometry.
Generally, the statistical information of points collected by the histograms can be categorized into spatial distributions and geometric attributes~\cite{Guo:2016:CPELFD}.
The former is adopted by descriptors like Spin Image~\cite{Johnson:1999:SPIN}, 3D Shape Context~\cite{Frome:2004:3DShapeContext}, and USC~\cite{Tombari:2010:USC:1877808.1877821};
while the latter is adopted by descriptors like PFH~\cite{Rusu:2008:PFH}, FPFH~\cite{Rusu:2009:FPFH}, and SHOT~\cite{Tombari:2010:USH,Salti:2014:SHOTUS}.
We refer the reader to a comprehensive survey by \Etal{Guo}~\shortcite{Guo:2016:CPELFD} on the hand-crafted descriptors.

\textbf{Learned 3D Local Descriptors.}
With the recent development of deep neural networks, significant research attention has focused on a data-driven approach to encode 3D local geometry into descriptors.
Existing works on learned descriptors generally differ in the choices of input parameterizations and network backbones.
To parameterize 3D local geometry, researchers have explored many representations such as 
voxel grids~\cite{Zeng_2017_CVPR,Gojcic_2019_CVPR,ao2020spinnet},
spherical signals~\cite{Spezialetti_2019_ICCV}, 
multi-view images~\cite{Huang:2017:LLS:3151031.3137609,Li_2020_CVPR},
radial histograms~\cite{Khoury_2017_ICCV},
and point pair features~\cite{Deng_2018_CVPR,Deng_2018_ECCV,Zhao_2019_CVPR}.
To extract descriptors, various network architectures have been leveraged, such as 
3D CNNs~\cite{Zeng_2017_CVPR,Gojcic_2019_CVPR},
Spherical CNNs~\cite{cohen2018spherical},
2D CNNs~\cite{Huang:2017:LLS:3151031.3137609,Li_2020_CVPR}, 
MLPs~\cite{Khoury_2017_ICCV}, 
PointNet~\cite{Deng_2018_CVPR,Deng_2018_ECCV,yew2018-3dfeatnet,Poiesi2021,Qi:2017:PointNetPlusPlus}, 
sparse convolutions~\cite{Choy_2019_ICCV,xie2020pointcontrast,Choy_2019_CVPR},
and kernel point convolutions~\cite{Bai_2020_CVPR,huang2020predator,Thomas_2019_ICCV}.
In this work, we base our descriptor extractor on 3DSmoothNet~\cite{Gojcic_2019_CVPR}, which uses a voxel-based representation and 3D CNNs.
Differently from that work, we propose a novel differentiable voxelization layer to enable descriptor extraction with a learnable local support instead of a predefined one, allowing the network to capture more representative local geometry in the descriptors.

Contrastive learning is normally adopted to optimize descriptor similarity.
For example, the triplet loss~\cite{Schroff_2015_CVPR,Hermans:2017:BH} used in 3DSmoothNet or the N-tuple~\cite{Deng_2018_CVPR,du2020dh3d} used in PPFNet.
To avoid the issues of ground-truth labeling, existing literature has further investigated unsupervised descriptor learning typically by taking auto-encoders~\cite{Yang_2018_CVPR} with a reconstruction loss~\cite{Deng_2018_ECCV,Zhao_2019_CVPR,Spezialetti_2019_ICCV}.
Closely related to the unsupervised approaches, our work proposes a novel registration loss, based on deviation from rigidity, to train the descriptor extractor without requiring ground-truth alignment information.

\textbf{Learning Geometric Registration.}
To learn local features well suited for registration, a myriad of studies have incorporated a differentiable registration layer into their networks, such as~\cite{yaoki2019pointnetlk,Wang_2019_ICCV,Wang:PRNet:2019,Deng_2019_CVPR,lawin2020registration,Yew_2020_CVPR,Gojcic_2020_CVPR,huang2020feature,choy2020deep,yan2021consistent}, among many others.
Works such as~\cite{choy2020deep,Pais_2020_CVPR,bai2021pointdsc,Yi_2018_CVPR} further examine feature learning of putative correspondences for outlier removal in pairwise matching.
Training the registration-based networks is normally done by minimizing the errors between the ground-truth and 3D transformations estimated by the networks.
Differently, our work zooms in informative local descriptor extraction and investigates a powerful registration-based training loss that penalizes deviation from rigidity for the estimated transformations, without requiring the knowledge of the ground-truth.

\section{Method}
\label{sec:Method}

As illustrated in \Figure{\ref{fig:pipeline}}, \OurMethodName{} takes a pair of 3D point clouds $\PointCloudP$ and $\PointCloudQ$ as input.
For a point $\PointP_{i} \in \PointCloudP$, to learn a robust local descriptor, our method has two stages: descriptor extraction and registration-based training.
In what follows, we first briefly discuss the pipeline of \OurMethodName{} and then present the details in the subsequent subsections.

To extract descriptors, we transform the local 3D  geometry into a voxel-based representation~\cite{Zeng_2017_CVPR,Gojcic_2019_CVPR} for the following reasons.
First, as an analogy to 2D images, the voxel-based representation is structured and works well with the off-the-shelf convolutional networks.
Second, as we show below, this allows us to optimize the local support size by directly making the voxel grid size a learnable parameter.
Third, unlike other local representations (\Eg, point pair features~\cite{Deng_2018_CVPR} and multi-view images~\cite{Li_2020_CVPR}) requiring additional information such as normals or colors, the voxel-based representation only depends on point coordinates, and is less sensitive to point density changes~\cite{Gojcic_2019_CVPR}.
Finally, using a local voxel-based representation can help to avoid overfitting to the global data modality (scene types) and thus offer better generalization ability than dense feature extractors (\Eg, KPConv~\cite{Bai_2020_CVPR}, PointConv~\cite{Wu_2019_CVPR}, and SparseConv~\cite{Choy_2019_ICCV}), as observed in existing works~\cite{Bai_2020_CVPR,ao2020spinnet,Poiesi2021}.

Our descriptor extraction network is built upon 3DSmoothNet~\cite{Gojcic_2019_CVPR}, which considers a fixed-size local neighborhood of $\PointP_{i}$ in voxelization.
We instead enable the network to learn the support size in a data-driven manner.
However, the conversion from point clouds to a voxel-based representation is a discrete operation lacking gradient definition.
To back-propagate the gradient to a learnable local support, we develop a differentiable voxelization layer based on probabilistic aggregation.
After the differentiable voxelization, we use a 3D CNN to extract robust features from the resulting voxel-based representation.

To train the descriptor extractor, we propose a weakly supervised registration loss that enforces rigidity in the pairwise alignment of point clouds with partial overlap.  
To formulate the registration loss, we first use the descriptors to build putative correspondences between point clouds.
Next, inspired by~\cite{roufosse2019unsupervised}, we propose to relax the alignment to be an affine transformation, which can be computed simply through weighted least-squares fitting.
This is different from existing end-to-end registration works~\cite{Wang:PRNet:2019,Wang_2019_ICCV} that directly solve for a rigid transformation via the SVD.
Finally, we define our registration loss by promoting structural properties of the computed transformation, including orthogonality and cycle consistency.

\subsection{Descriptor Extraction}
\label{subsec:DescriptorExtraction}

\begin{figure}[h]
	\centering
	\includegraphics[width=\linewidth]{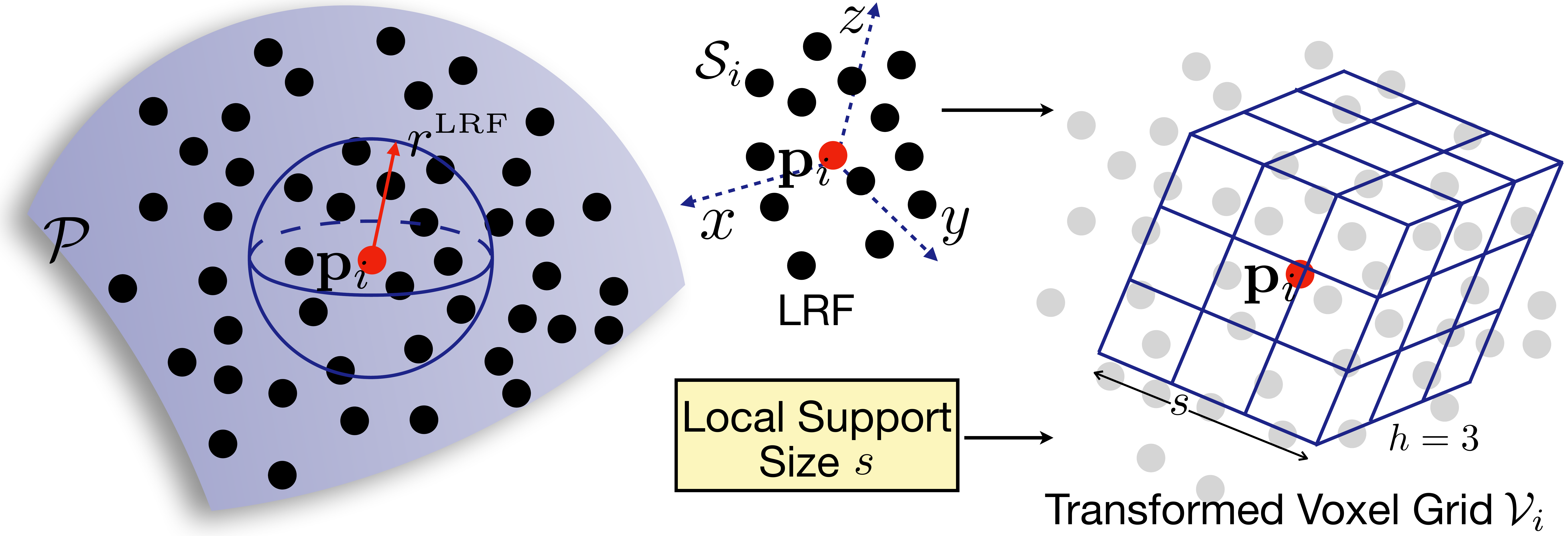}
	\caption{Illustration of transforming the local geometry of a point $\PointP_{i}$ to a voxel-based representation $\VoxelGrid_{i}$. The resolution of $\VoxelGrid_{i}$ is $\VoxelGridResolution^{3}$, and the voxel size is $\VoxelGridScale / \VoxelGridResolution$.}
	\label{fig:input_parameterization}
\end{figure}

To extract a robust descriptor for the point $\PointP_{i}$, we first convert its local geometry in $\PointCloudP$ to a function defined over a voxel grid $\VoxelGrid_{i}$ (\Figure{\ref{fig:input_parameterization}}).
Suppose that $\VoxelGrid_{i}$ has a resolution of $\VoxelGridResolution \times \VoxelGridResolution \times \VoxelGridResolution$, where $\VoxelGridResolution$ is an integer hyperparameter.
Prior to voxelization, we need to determine the orientation and the size of the local support of $\PointP_{i}$.
The orientation is for ensuring rotation invariance, and the support size determines the amount of local geometry information captured in the final descriptor.

For the orientation, as discussed in~\cite{Gojcic_2019_CVPR}, it is nontrivial to perform its regression as an integral part of neural networks~\cite{Deng_2018_CVPR,Esteves_2018_ECCV,spezialetti2020learning}.
Instead, we use the approach based on explicit local reference frame (LRF) estimation~\cite{YANG2017175} in 3DSmoothNet.
Specifically, a local patch $\Neighborhood_{i} \subset \PointCloudP$ centered at $\PointP_{i}$ is extracted (\Figure{\ref{fig:input_parameterization}}).
The local patch is defined as $\Neighborhood_{i} = \{ \PointP_{j} \ | \ \| \PointP_{j} - \PointP_{i} \|_2 \leq \NeighborhoodRadius \}$, where $\NeighborhoodRadius$ is a predefined radius.
We follow~\cite{Gojcic_2019_CVPR} to compute the LRF with resolved sign ambiguity based on eigendecomposition of the covariance matrix of the points in $\Neighborhood_{i}$.
We stack the axes of the resulting LRF as column vectors in a matrix $\LRFTransform_{i} \in \mathbb{R}^{3\times3}$.

For the support size, 3DSmoothNet empirically uses a fixed setting $\VoxelGridScale = \frac{2\NeighborhoodRadius}{\sqrt{3}}$ as the grid size of $\VoxelGrid_{i}$ for enclosing $\Neighborhood_{i}$.
In contrast, we integrate the voxelization into the network and enable the support size $\VoxelGridScale$ to be a learnable parameter during training.
In this way, the network can gain better flexibility to capture representative local geometry not circumscribed by the predefined  $\Neighborhood_{i}$, thus boosting the geometric informativeness in the learned descriptors (\Section{\ref{subsec:3DMatchDataset}}).

Next, as shown in \Figure{\ref{fig:input_parameterization}} (right), we anchor the center of the voxel grid $\VoxelGrid_{i}$ at $\PointP_{i}$, and rotate $\VoxelGrid_{i}$ by $\LRFTransform_{i}$ for alignment with the LRF. 
The grid size of $\VoxelGrid_{i}$ is set to the learnable parameter $\VoxelGridScale$.

\textbf{Differentiable Voxelization.}
The conventional voxelization is non-differentiable and thus cannot back-propagate gradient \Wrt{} training losses to the local support size optimization.
To address this, we propose point-to-voxel conversion within the network and design a differentiable voxelization layer leveraging probabilistic aggregation~\cite{Liu_2019_ICCV,Li_2020_CVPR}.

\begin{figure}[h]
	\centering
	\includegraphics[width=\linewidth]{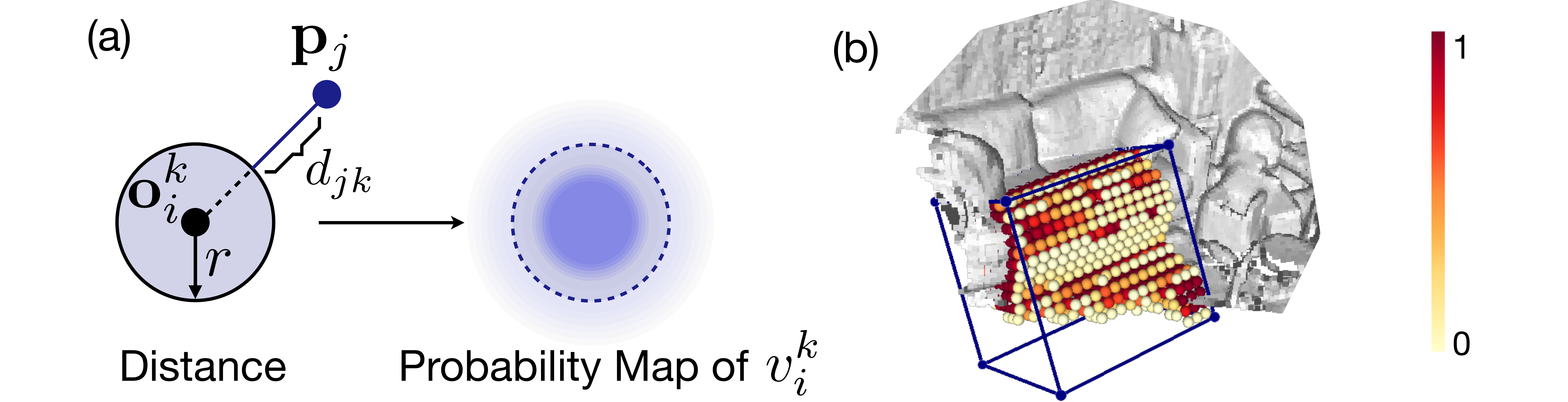}
	\caption{Illustration of the differentiable voxelization layer. (a) Probability map computation for voxel $\Voxel_{i}^{k} \in \VoxelGrid_{i}$. (b) Voxelization example; for clarity, only voxels with values $\ge 0.01$ are visualized.}
	\label{fig:differentiable_voxelization}
\end{figure}

For the $k$-th voxel $\Voxel_{i}^{k}$ in the transformed $\VoxelGrid_{i}$, we compute its value in a probabilistic manner.
To simplify computation, we consider each voxel as a sphere with a radius of 
$\VoxelRadius = \frac{\VoxelGridScale}{2\VoxelGridResolution}$~\cite{Qi_2016_CVPR}.
We use $\VoxelPointProbability_{jk}$ to denote the probability of some point $\PointP_j \in \PointCloudP$ contained in the  voxel $\Voxel_{i}^{k}$ (\Figure{\ref{fig:differentiable_voxelization}-a}):
\begin{equation}
	\label{eq:VoxelPointProbability}
	\VoxelPointProbability_{jk} = \text{sigmoid}(\delta_{jk} \cdot \dfrac{\VoxelPointDistance_{jk}^2}{\sigma}), \ \VoxelPointDistance_{jk} = \| \PointP_j - \VoxelCenter_{i}^{k} \|_2 - \VoxelRadius,
\end{equation}
where $\VoxelCenter_{i}^{k} \in \mathbb{R}^3$ denotes the voxel center, $\delta_{jk}$ is a sign indicator $\delta_{jk} = - \textrm{sign} (\VoxelPointDistance_{jk})$, and $\sigma$ controls the sharpness of the probability distribution.
We then propose to aggregate the voxel value of $\Voxel_{i}^{k}$ as:
\begin{equation}
	\label{eq:DifferentiableVoxelization}
	\Voxel_{i}^{k} = 1 - \prod_{\PointP_j \in \PointCloudP} (1 - \VoxelPointProbability_{jk} ).
\end{equation}
In \Equation{\eqref{eq:DifferentiableVoxelization}}, the voxel value can be viewed as the probability of having at least one point contained in the voxel $\Voxel_{i}^{k}$. 
The resulting voxel values are a continuous function of the point cloud coordinates and of the learnable support size $\VoxelGridScale$, and thus \Equation{\eqref{eq:DifferentiableVoxelization}} is fully differentiable.
We also note that \Equation{\eqref{eq:DifferentiableVoxelization}} is permutation invariant to the input points and less sensitive to point density changes, thus accommodating an arbitrary number of points.
In \Figure{\ref{fig:differentiable_voxelization}-b}, we visualize a voxelization example, showing that probabilistic aggregation can approximate the voxelization procedure to capture the structure of local geometry.

\begin{figure}[h]
	\centering
	\includegraphics[width=\linewidth]{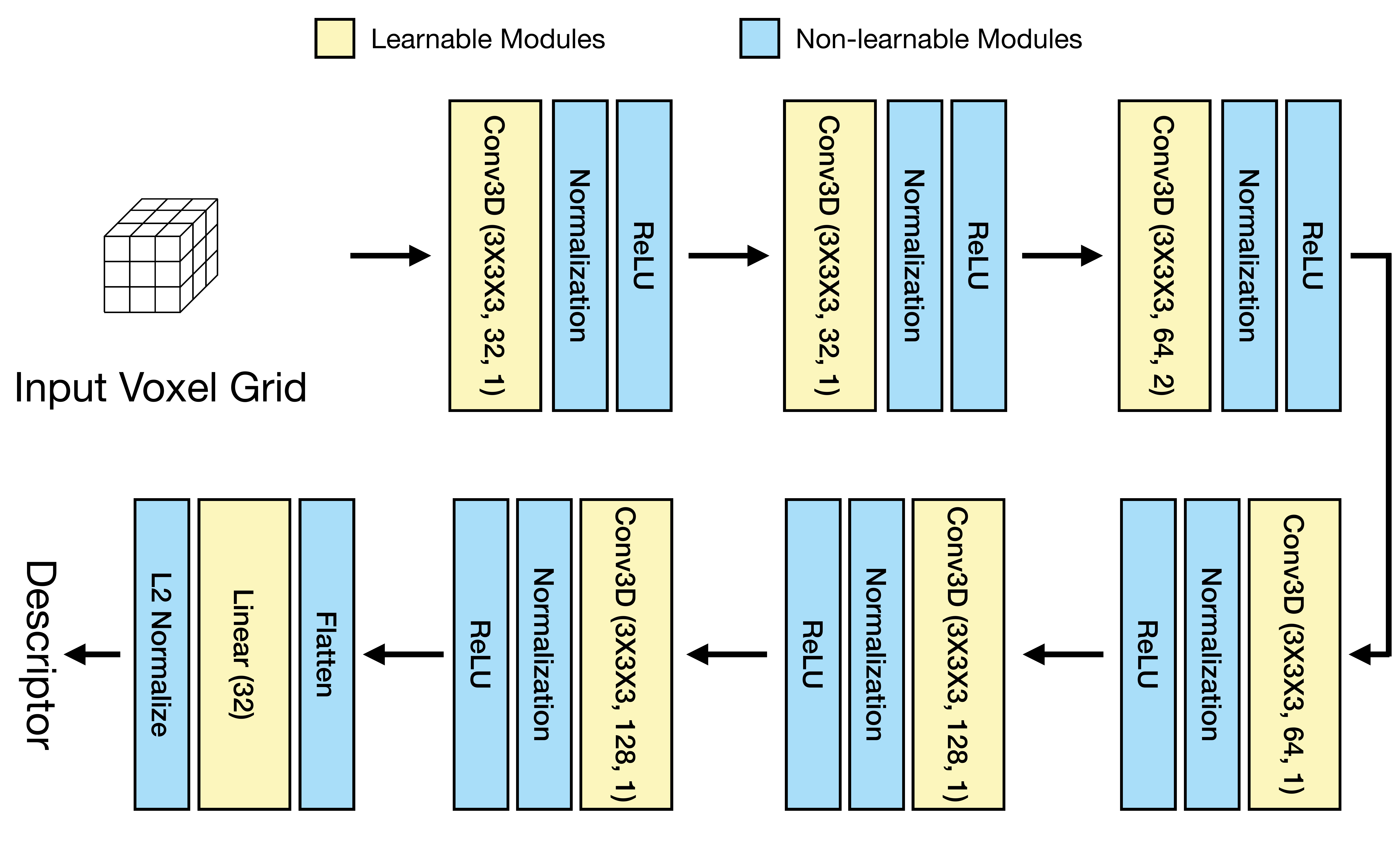}
	\caption{The 3D CNN-based descriptor extractor. The numbers in the parentheses of each 3D convolution layer represent kernel size, number of filters, and stride, respectively. The linear layer outputs $\DescriptorDimension$-dimensional descriptors.}
	\label{fig:conv3d_architecture}
\end{figure}

\textbf{3D CNN.}
After converting the local geometry of $\PointP_{i}$ to a signal on the voxel grid $\VoxelGrid_{i}$, we use the 3D CNN-based architecture from~\cite{Gojcic_2019_CVPR} to compute a descriptor $\Descriptor_{i} \in \mathbb{R}^{\DescriptorDimension}$.
Let $\Descriptor_{i} = \ConvNet_{\ConvNetParams} (\VoxelGrid_{i})$, where $\ConvNetParams$ denotes the learnable parameters of the 3D CNN $\ConvNet$.
\Figure{\ref{fig:conv3d_architecture}} presents the details of $\ConvNet$.
Specifically, the network $\ConvNet$ is comprised of six convolutional layers, which
progressively down-sample the input $\VoxelGrid_{i}$.
Each convolutional layer is followed by a normalization layer~\cite{UlyanovVL16:InstNorm} and a ReLU activation layer.
The output of the last convolutional layer is flattened and fed to a fully connected layer followed by $\ell^2$ normalization, resulting a unit-length $\DescriptorDimension$-dimensional local descriptor $\Descriptor_{i}$.

\subsection{Registration-based Training}
\label{subsec:UnsupervisedTraining}

In this section, we propose a novel registration loss for optimizing descriptor similarity across point clouds.
To formulate the loss, we assume the input point clouds $\PointCloudP$ and $\PointCloudQ$ to have partial overlap.
Specifically, partial overlap means the existence of corresponding points between $\PointCloudP$ and $\PointCloudQ$, and the overlap ratio~\cite{Choi_2015_CVPR} is normally defined as the number of corresponding points over $\min\{|\PointCloudP|, |\PointCloudQ|\}$.
Having partial overlap is a very weak supervision for ensuring the \emph{presence} of some underlying rigid alignment.
Note that this weak supervision can be easily satisfied in existing 3D point cloud datasets~\cite{Zeng_2017_CVPR}, such as by leveraging temporally adjacent point clouds due to the continuity of the scanning process in practice or by employing a RANSAC-based overlap detection procedure used in the reconstruction pipelines~\cite{zhou2016fast,Zhou:2018:Open3D}.
We stress that our loss calculation \emph{does not} require the knowledge of the ground-truth transformation between $\PointCloudP$ and $\PointCloudQ$.

\textbf{Putative Correspondences.}
We construct a set of putative correspondences between $\PointCloudP$ and $\PointCloudQ$ by a nearest neighbor (NN) search between their extracted local descriptors.
Note that the hard assignments of the NN search are not differentiable.
Thus we adopt a differentiable NN search based on the softmin operation, which has been used in the prior works~\cite{Gojcic_2020_CVPR,plotz2018neural}.
We use $\CorrespondenceSet$ to denote the resulting set of correspondences $\{ \Correspondence_{i} = (\PointP_{i} \in \PointCloudP, \PointQ_{i} \in \PointCloudQ) \}$,  where for a point $\PointP_{i}$ its closest neighbor in the descriptor space is the point $\PointQ_{i}$.

\textbf{Transformation Estimation.}
Given the constructed $\CorrespondenceSet$, we solve for a 3D transformation aligning $\PointCloudP$ to $\PointCloudQ$ and then formulate our registration loss by promoting structural properties of the resulting transformation.
We denote the transformation as $(\PointCloudRotation, \PointCloudTranslation)$, where $\PointCloudRotation$ is a $3 \times 3$ affine matrix and $\PointCloudTranslation$ is a 3D vector.
We stack the points $\{\PointP_{i}\}$ in $\CorrespondenceSet$ as columns of a matrix $\PointCloudPMat_{c} \in \mathbb{R}^{3 \times |\CorrespondenceSet|}$, and $\PointCloudQMat_{c}$ is a similar matrix stacking the points $\{\PointQ_{i}\}$ in $\CorrespondenceSet$.
To estimate $(\PointCloudRotation, \PointCloudTranslation)$, we minimize the following weighted quadratic error:
\begin{equation}
	\label{eq:LeastSquaresEnergy}
	\PointCloudRotation, \PointCloudTranslation = \ArgMin_{\PointCloudRotation, \PointCloudTranslation} \|  ( \PointCloudRotation \PointCloudPMat_{c} + \PointCloudTranslation \mathbf{1} - \PointCloudQMat_{c} ) \CorrespondenceWeightMat \|^{2},
\end{equation}
where $\mathbf{1}$ is a row vector filled with 1, and the matrix $\CorrespondenceWeightMat$ denotes the correspondence confidence: $\CorrespondenceWeightMat = \text{diag} (\CorrespondenceWeightEntry_{1}, \cdots, \CorrespondenceWeightEntry_{i}, \cdots, \CorrespondenceWeightEntry_{|\CorrespondenceSet|})$ computed as follows.

For the $i$-th correspondence, we define its confidence as
\begin{equation}
    \label{eq:CorrespondenceWeight}
    \CorrespondenceWeightEntry_{i} = \CorrespondenceWeightEntry_{i}^{f} \cdot \CorrespondenceWeightEntry_{i}^{sm},
\end{equation}
where $\CorrespondenceWeightEntry_{i}^{f}$ is the similarity of the two points in the descriptor space, and $\CorrespondenceWeightEntry_{i}^{sm}$ is the compatibility with other correspondences in the 3D space.
Specifically, given the $i$-th correspondence $\Correspondence_{i} = (\PointP_{i}, \PointQ_{i})$ with local descriptors $\Descriptor_{\PointP_{i}}$ and $\Descriptor_{\PointQ_{i}}$, 
the first term $\CorrespondenceWeightEntry_{i}^{f}$ is computed by the softmax function: 
\begin{equation}
	\CorrespondenceWeightEntry_{i}^{f} = \dfrac{\exp(- \| \Descriptor_{\PointP_{i}} - \Descriptor_{\PointQ_{i}} \|_2)}{\sum_{\PointQ_{j} \in \PointCloudQ} \exp(- \| \Descriptor_{\PointP_{i}} - \Descriptor_{\PointQ_{j}} \|_2)}.
\end{equation}
For the second term $\CorrespondenceWeightEntry_{i}^{sm}$, we follow~\cite{bai2021pointdsc} and leverage the well-known spectral matching technique~\cite{leordeanu2005spectral}, which finds consistent correspondences based on their isometric compatibility.
Spectral matching does not require ground-truth labels and is fully differentiable.
Details of this technique can be found in the supplementary material and in~\cite{leordeanu2005spectral,golub1996matrix}.

We note that while there exist recent deep outlier filtering works~\cite{Yi_2018_CVPR,choy2020deep,Pais_2020_CVPR} using neural networks to regress correspondence confidence, we adopt \Equation{\eqref{eq:CorrespondenceWeight}} for the following reasons.
First, those deep outlier filtering methods require to be fully supervised for the regression, and thus cannot be adapted straightforwardly in our work, which strives to use very weak supervision for local descriptor learning.
Second, the networks used in the above works typically have a myriad of trainable parameters, which can significantly increase the complexity of our network. 
In contrast, the above computation for $\CorrespondenceWeightMat$ is non-parametric and fully differentiable, allowing gradient back-propagation to the descriptor extractor.
At test time, following 3DSmoothNet, we combine our extracted local descriptors with RANSAC~\cite{Fischler:1981:RANSAC} for robust point cloud registration.

\textbf{Registration Loss.}
We formulate our registration loss based on the transformation estimation by \Equation{\eqref{eq:LeastSquaresEnergy}}.
Prior registration-based works~\cite{Wang_2019_ICCV,Wang:PRNet:2019,Yew_2020_CVPR,Gojcic_2020_CVPR} employ the SVD to solve a similar optimization problem, ensuring that $(\PointCloudRotation, \PointCloudTranslation)$ is a rigid transformation that can be compared with the ground-truth during training.
Differently, we propose to relax $(\PointCloudRotation, \PointCloudTranslation)$ to be an affine transformation and leverage its \emph{deviation} from rigidity as the driving force for local descriptor training, as done in the context of non-rigid shape matching~\cite{roufosse2019unsupervised}.
Specifically, with the affine relaxation, we can solve \Equation{\eqref{eq:LeastSquaresEnergy}} in a least-squares sense as follows:
\begin{equation}
	\label{eq:LeastSquaresEnergySolution}
	\big[ \PointCloudRotation \ \PointCloudTranslation \big] = \PointCloudQMat_{c} \CorrespondenceWeightMat (\overline{\PointCloudPMat}_{c} \CorrespondenceWeightMat)^{+},
\end{equation}
where $\overline{\PointCloudPMat}_{c} \in \mathbb{R}^{4 \times |\CorrespondenceSet|}$ is $\PointCloudPMat_{c}$ in homogeneous coordinates, and $(\cdot)^{+}$ is the Moore–Penrose inverse, which is differentiable.
The inverse transformation $(\PointCloudRotationInv,  \PointCloudTranslationInv)$ aligning $\PointCloudQ$ to $\PointCloudP$ is computed similarly.

We denote our registration loss as $\LossRegistration$, consisting of the following terms:
\begin{align}
	\LossOrthogonality &= \big(\| \PointCloudRotation^{\top} \PointCloudRotation - \IdentityMatrix \|_{1} + \| \PointCloudRotationInv^{\top} \PointCloudRotationInv - \IdentityMatrix \|_{1} \big) / 2, \label{eq:LossOrthogonality}\\
	\LossCycle &= \| \PointCloudRotation \PointCloudRotationInv - \IdentityMatrix \|_{1} + \| \PointCloudRotation \PointCloudTranslationInv + \PointCloudTranslation \|_{1}, \label{eq:LossCycle}\\
	\LossRegistration &= \LossOrthogonalityWeight \LossOrthogonality + \LossCycleWeight \LossCycle,\label{eq:LossRegistration}
\end{align}
where $\LossOrthogonalityWeight$ and $\LossCycleWeight$ are the weights for the loss terms, and $\IdentityMatrix$ is an identity matrix.
The term $\LossOrthogonality$ regularizes the orthogonality of $\PointCloudRotation$ and $\PointCloudRotationInv$. The term $\LossCycle$ enforces the cycle consistency
by requiring 
\begin{equation}
	\label{eq:CycleConsistency}
	\begin{bmatrix} \PointCloudRotation & \PointCloudTranslation \\ \mathbf{0} & 1 \end{bmatrix} \begin{bmatrix} \PointCloudRotationInv & \PointCloudTranslationInv \\ \mathbf{0} & 1 \end{bmatrix} = \begin{bmatrix} \PointCloudRotation \PointCloudRotationInv &  \PointCloudRotation \PointCloudTranslationInv + \PointCloudTranslation \\ \mathbf{0} & 1 \end{bmatrix} = \IdentityMatrix .
\end{equation}

\textbf{Discussion.}
We briefly discuss the intuition behind $\LossRegistration$ for providing training signals without requiring ground-truth alignment information.
If $\CorrespondenceSet$ is a set of inlier correspondences, the underlying rigid transformation can always be recovered by \Equation{\eqref{eq:LeastSquaresEnergySolution}}.
However, outlier correspondences may exist in $\CorrespondenceSet$ during the local descriptor training, making the transformation by \Equation{\eqref{eq:LeastSquaresEnergySolution}} deviate from rigidity.
Minimizing this deviation in $\LossRegistration$ will translate to a tendency of promoting weights for the inliers and lowering weights for the outliers (\Equation{\eqref{eq:LeastSquaresEnergy}}).
The gradient \Wrt{} $\LossRegistration$ can be back-propagated through the differentiable correspondence matching layer to the descriptor extractor, thus enabling the similarity optimization between local descriptors.

\subsection{Implementation}
\label{subsec:Implementation}

We implemented our method with PyTorch~\cite{Pytorch:NIPS2019}.
Following 3DSmoothNet~\cite{Gojcic_2019_CVPR}, we set $\NeighborhoodRadius = 0.3$, the voxel grid resolution $\VoxelGridResolution = 16$, and the descriptor dimension $\DescriptorDimension = 32$.
In the differentiable voxelization, we set $\sigma=10^{-3}$.
For the term weights in $\LossRegistration$, we use $\LossOrthogonalityWeight = 1$ and $\LossCycleWeight = 1$.
We adopt Adam~\cite{KingmaB14:Adam} with a learning rate of $10^{-3}$ as the network optimizer.
We use two point cloud datasets, 3DMatch~\cite{Zeng_2017_CVPR} and ModelNet40~\cite{Wu:2015:3DSN}.
During training on the 3DMatch dataset, we sample 512 keypoints in each point cloud with farthest point sampling for sparse descriptor extraction and matching.
The network is trained for 16K steps.
For the ModelNet40 dataset, we sample 128 keypoints in each point cloud during training.
The network is trained for 20K steps.

\section{Experiments}
\label{sec:Experiments}

We evaluate the performance of our proposed \OurMethodName{} on existing point cloud registration benchmarks including 3DMatch (\Section{\ref{subsec:3DMatchDataset}}), ModelNet40 (\Section{\ref{subsec:ModelNet40Dataset}}), and ETH (\Section{\ref{subsec:ETHDataset}}).
The 3DMatch dataset~\cite{Zeng_2017_CVPR} consists of point clouds of indoor scene scans.
The ModelNet40 dataset~\cite{Wu:2015:3DSN} consists of object-centric point clouds generated from CAD models.
The ETH dataset~\cite{Pomerleau:2012} consists of outdoor scene scans and is used for descriptor generalization test.
We present the ablation study in \Section{\ref{subsec:AblationStudy}}.

\subsection{3DMatch Dataset}
\label{subsec:3DMatchDataset}
The 3DMatch dataset is widely adopted for evaluating the local descriptor performance on geometric registration.
In total, there are 62 indoor scenes: 54 of them for training and validation, and 8 of them for testing.
In the test set, the number of points per point cloud is $\sim$13K on average, and each point cloud has 5K randomly sampled keypoints for sparse descriptor extraction and matching.

\textbf{Evaluation Metrics.}
Following~\cite{Gojcic_2019_CVPR,Choy_2019_ICCV}, we compute the inlier ratio (IR), feature-match recall (FMR), and registration recall (RR) on 3DMatch.
Consider a set of testing point cloud pairs $\PointCloudPairSet = \{ (\PointCloudP, \PointCloudQ) \}$ with $\PointCloudP$ and $\PointCloudQ$ having at least 30\% overlap.
For each point cloud pair, a set of putative correspondences $\PutativeCorrespondenceSet$ between keypoints is built by finding mutually closest neighbors in the descriptor space:
\begin{equation}
\label{suppeq:PutativeCorrespondenceSet}
\resizebox{0.85\columnwidth}{!}{%
	$\begin{split}
		\PutativeCorrespondenceSet = 
		\{
		(\PointP_{i} \in \PointCloudP, \PointQ_{i} \in \PointCloudQ) | 
		\DescriptorExtraction(\PointP_{i}) = \text{NN}(\DescriptorExtraction(\PointQ_{i}), \DescriptorExtraction(\PointCloudP)) \wedge \\
		\DescriptorExtraction(\PointQ_{i}) = \text{NN}(\DescriptorExtraction(\PointP_{i}), \DescriptorExtraction(\PointCloudQ))
		\},
	\end{split}$
}
\end{equation}
where $\DescriptorExtraction$ denotes a specific descriptor extractor, and $\text{NN} (\cdot, \cdot)$ is the nearest neighbor query based on the $\ell^2$ distance.

IR measures the fraction of correct correspondences between $\PointCloudP$ and $\PointCloudQ$ as follows:
\begin{equation}
	\label{suppeq:IR}
	\text{IR} = \dfrac{1}{|\PutativeCorrespondenceSet|} \sum_{(\PointP_{i}, \PointQ_{i}) \in \PutativeCorrespondenceSet} 
	\Big\llbracket
	\| \PointCloudTransform^{*} (\PointP_{i}) - \PointQ_{i} \|_{2} < \InlierThresh
	\Big\rrbracket,
\end{equation}
where $\PointCloudTransform^{*}(\cdot)$ denotes the ground-truth transformation,
$\InlierThresh = 0.1 \text{ m}$ is the inlier distance threshold, and $\llbracket \cdot \rrbracket$ is the Iverson bracket.

FMR measures the fraction of point cloud pairs in $\PointCloudPairSet$, for which a RANSAC-based~\cite{Fischler:1981:RANSAC} registration pipeline can recover the transformations with high confidence~\cite{Deng_2018_CVPR}:
\begin{equation}
	\text{FMR}= \dfrac{1}{|\PointCloudPairSet|} \sum_{(\PointCloudP, \PointCloudQ) \in \PointCloudPairSet} \big\llbracket \text{IR} > \InlierRatioThresh  \big\rrbracket,
\end{equation}
where $\InlierRatioThresh$ is the inlier ratio threshold in the range of $[0.05, 0.2]$.

RR examines the performance of local descriptors in an actual reconstruction system.
For a pair of point clouds $\PointCloudP$ and $\PointCloudQ$, let $\hat{\PointCloudTransform}(\cdot)$ denote the estimated transformation by a registration pipeline.
Suppose $\GTCorrespondenceSet$ is the set of ground-truth correspondences, and their root-mean-square error is computed as follows:
\begin{equation}
	\label{suppeq:RMSE}
	\text{RMSE} = \sqrt{\dfrac{1}{|\GTCorrespondenceSet|} \sum_{(\PointP_{i}, \PointQ_{i}) \in \GTCorrespondenceSet} \| \hat{\PointCloudTransform}(\PointP_{i}) - \PointQ_{i} \|_{2}^{2}}.
\end{equation} 
RR is computed as the fraction of point cloud pairs in $\PointCloudPairSet$ with $\text{RMSE} < 0.2 \text{ m}$.

\textbf{Comparisons.}
We compare our descriptor with hand-crafted descriptors and existing descriptor learning methods that do not rely on ground-truth transformations.
The former includes FPFH~\cite{Rusu:2009:FPFH} and SHOT~\cite{Salti:2014:SHOTUS}, which have been implemented in the PCL library~\cite{Rusu:2011:PCL} and have descriptor dimensions of 33 and 352, respectively.
The latter includes PPF-FoldNet~\cite{Deng_2018_ECCV}, CapsuleNet~\cite{Zhao_2019_CVPR}, and S2CNN~\cite{Spezialetti_2019_ICCV}.
Their implementations are based on publicly available codebases\footnote{\url{https://github.com/XuyangBai/PPF-FoldNet}}\footnote{\url{https://github.com/yongheng1991/3D-point-capsule-networks}}\footnote{\url{https://github.com/jonas-koehler/s2cnn}}, 
and they all have a descriptor dimension of 512.
The high dimensionality makes NN search computationally inefficient.
Thus, for a direct comparison with the state-of-the-art S2CNN, we also implemented an S2CNN variant that outputs 32-dimensional descriptors.

\begin{table}[h]
\centering
\caption{Performance (\%) on the 3DMatch dataset \Wrt{} inlier ratio (IR), feature-match recall (FMR), and registration recall (RR). The best and the second best results among the methods w/o GT are highlighted. 
(GT -- ground-truth labels; Dim. -- descriptor dimension.)}
\resizebox{\columnwidth}{!}{%
\begin{tabular}{l@{\hspace{0.1em}}|c|ccccc}
\toprule
                                                & GT                 & Dim.          & IR$\uparrow$       & \multicolumn{2}{c}{FMR$\uparrow$}       & RR$\uparrow$        \\
$\InlierRatioThresh$                            &                    &               &                    & 0.05                & 0.2                &                     \\
\midrule                                                                                                                                                                       
3DMatch                                         & w/                 &         512   &               8.3  &            57.3     &             7.7    &               51.9  \\
CGF                                             & w/                 &         32    &              10.1  &            60.6     &            12.3    &               51.3  \\
PPFNet                                          & w/                 &         64    &              -     &            62.3     &            -       &               71.0  \\
3DSmoothNet                                     & w/                 &         32    &              36.0  &            95.0     &            72.9    &               78.8  \\
FCGF                                            & w/                 &         32    &              -     &            95.2     &            67.4    &               82.0  \\
D3Feat                                          & w/                 &         32    &              40.7  &            95.8     &            75.8    &               82.2  \\
LMVD                                            & w/                 &         32    &              46.1  &            97.5     &            86.9    &               81.3  \\
SpinNet                                         & w/                 &         32    &              -     &            97.6     &            85.7    &               -     \\
\OurMethodName{} (BH)                           & w/                 &         32    &              54.3  &            96.7     &            88.2    &               81.4  \\
\midrule                                                                                                                                                                        
FPFH                                            & w/o                &         33    &               9.3  &            59.6     &            10.1    &               54.8  \\
SHOT                                            & w/o                &         352   &              14.9  &            73.3     &            26.9    &               59.4  \\
PPF-FoldNet                                     & w/o                &         512   &              20.9  &            83.8     &            41.0    &               69.0  \\
CapsuleNet                                      & w/o                &         512   &              16.9  &            82.5     &            31.3    &               67.6  \\
S2CNN                                           & w/o                &         512   &   \underline{34.5} & \underline{94.6}    & \underline{70.3}   &    \underline{78.4} \\
S2CNN                                           & w/o                &         32    &              29.1  &            92.4     &            59.9    &               73.9  \\
\OurMethodName                                  & w/o                &         32    &      \textbf{42.3} &    \textbf{95.1}    &    \textbf{77.7}   &       \textbf{80.0} \\
\bottomrule
\end{tabular}
}
\label{tab:3dmatch_fmr_ir_rr}
\end{table}

\begin{figure}[h]
	\centering
	\includegraphics[width=\linewidth]{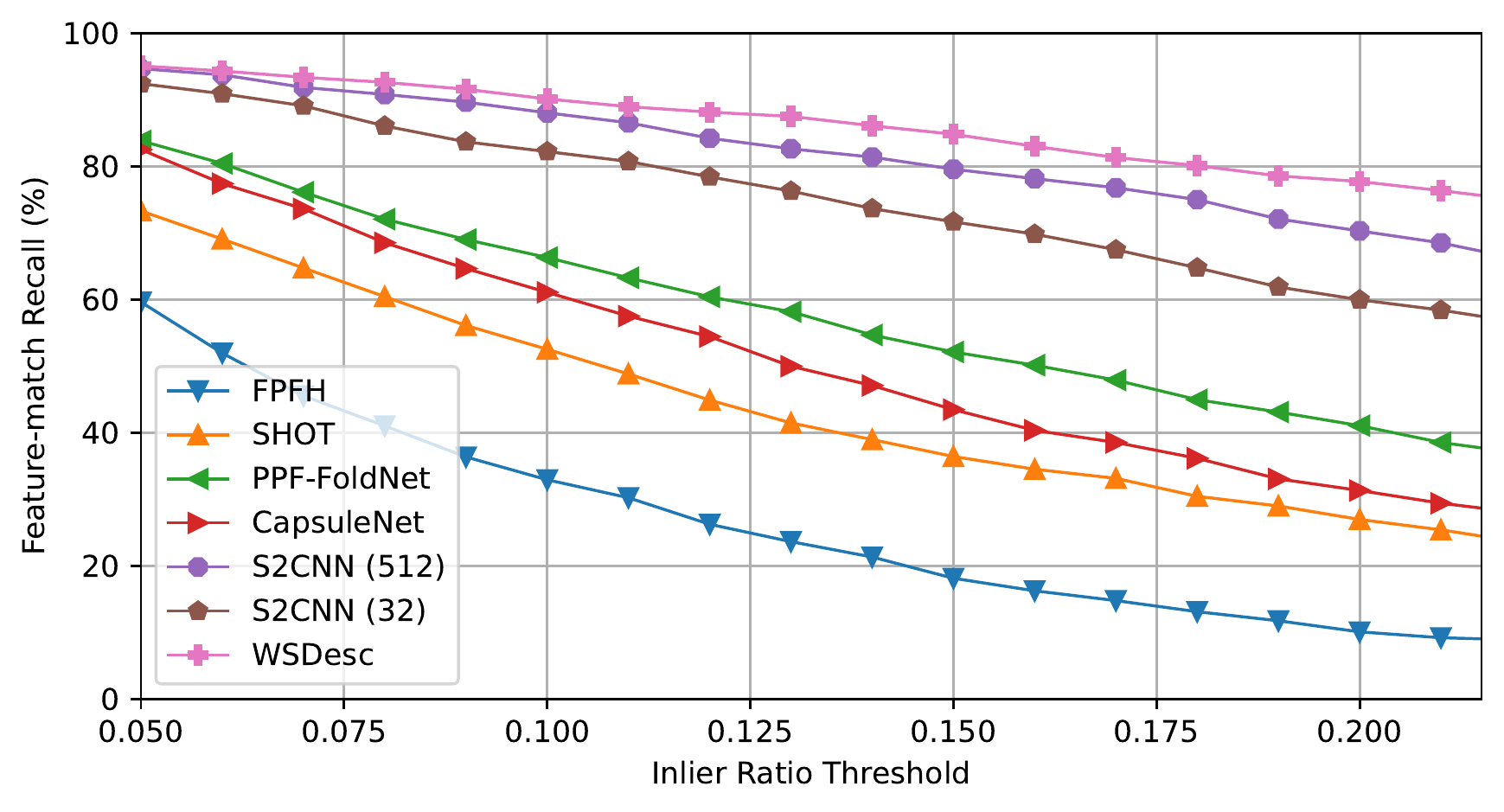}
	\caption{Feature-match recall of the methods w/o GT on 3DMatch, with a varying inlier ratio threshold $\InlierRatioThresh$.}
	\label{fig:3dmatch_fmr_varying_tau2}
\end{figure}

\Table{\ref{tab:3dmatch_fmr_ir_rr}} (bottom) shows the IR comparison for the methods w/o GT.
It can be observed that our descriptor obtains the highest IR (42.3\%).
\OurMethodName{} outperforms S2CNN (32) by 13.2 percentage points and S2CNN (512) by 7.8 percentage points, indicating the better quality of correspondences built by our method.
For the FMR comparison, in the case of $\InlierRatioThresh = 0.05$, our descriptor achieves an FMR of 95.1\%, slightly better than S2CNN (512).
Yet $\InlierRatioThresh = 0.05$ is a relatively easy threshold~\cite{Gojcic_2019_CVPR}, and the descriptor performance tends to be saturated.
In the harder case of $\InlierRatioThresh = 0.2$, our descriptor obtains 77.7\%, significantly outperforming S2CNN (32) by 17.8 percentage points and S2CNN (512) by 7.4 percentage points.
Further, \Figure{\ref{fig:3dmatch_fmr_varying_tau2}} plots the FMR performance \Wrt{} different $\InlierRatioThresh$ values, and our descriptor shows lower sensitivity to the inlier ratio $\InlierRatioThresh$, which can be ascribed to the superior IR performance.
For the RR metric, our descriptor also achieves the best performance (80.0\%) and outperforms S2CNN (32) by 6.1 percentage points.

In \Table{\ref{tab:3dmatch_fmr_ir_rr}} (top) we include the performance of supervised descriptor learning methods, including 3DMatch~\cite{Zeng_2017_CVPR}, CGF~\cite{Khoury_2017_ICCV}, PPFNet~\cite{Deng_2018_CVPR}, 3DSmoothNet~\cite{Gojcic_2019_CVPR}, FCGF~\cite{Choy_2019_ICCV}, D3Feat~\cite{Bai_2020_CVPR}, LMVD~\cite{Li_2020_CVPR}, and SpinNet~\cite{ao2020spinnet}.
Interestingly, our \OurMethodName{} achieves even better performance than the supervised 3DSmoothNet, 
which can be ascribed mainly to our learnable support that allows capturing the local context in an appropriate size.
To validate this, \OurMethodName{} (BH) in \Table{\ref{tab:3dmatch_fmr_ir_rr}} is the oracle performance of our method, that is, we use the same supervised batch-hard loss as 3DSmoothNet for training instead of $\LossRegistration$.
It can be found that using the learnable support significantly improves the performance of 3DSmoothNet, making it comparable to the state-of-the-arts~\cite{Bai_2020_CVPR,Li_2020_CVPR,ao2020spinnet}.

\textbf{Rotation Invariance.}
To test the rotation invariance, following~\cite{Deng_2018_ECCV}, we used the rotated 3DMatch dataset, where each point cloud is rotated with randomly sampled axes and angles in $[0, 2\pi]$.
\Table{\ref{tab:3dmatch_rotated_fmr_ir_rr}} reports the performance of the compared methods.
Our descriptor has the best IR (39.8\%), FMR (74.3\% at $\InlierRatioThresh = 0.2$), and RR (78.5\%) scores.

\begin{table}[h]
\centering
\caption{Performance (\%) on the rotated 3DMatch dataset. $\InlierRatioThresh = 0.2$ for FMR.}
\resizebox{0.75\columnwidth}{!}{%
\begin{tabular}{l|cccc}
\toprule
                                                & Dim.          & IR$\uparrow$       & FMR$\uparrow$       & RR$\uparrow$        \\
\midrule                                                                                                                                                                       
FPFH                                            &         33    &               9.3  &            10.0     &               55.3  \\
SHOT                                            &         352   &              14.9  &            26.9     &               61.6  \\
PPF-FoldNet                                     &         512   &              21.0  &            41.6     &               68.7  \\
CapsuleNet                                      &         512   &              16.8  &            31.9     &               68.0  \\
S2CNN                                           &         512   &   \underline{34.6} & \underline{70.5}    &    \underline{78.3} \\
S2CNN                                           &         32    &              29.2  &            59.5     &               75.2  \\
\OurMethodName                                  &         32    &      \textbf{39.8} &    \textbf{74.3}    &       \textbf{78.5} \\
\bottomrule
\end{tabular}
}
\label{tab:3dmatch_rotated_fmr_ir_rr}
\end{table}

\begin{figure*}[h]
	\centering
	\includegraphics[width=\linewidth]{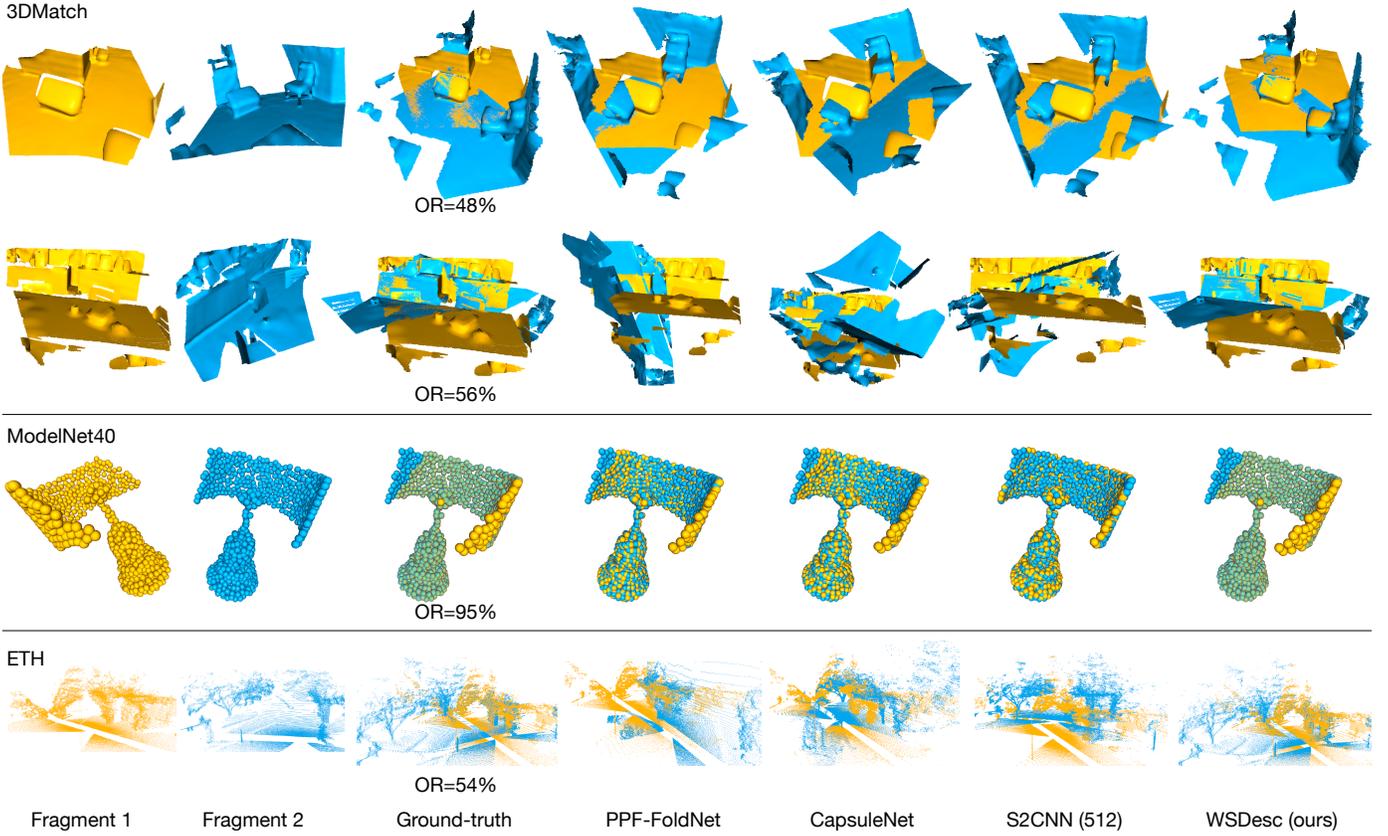}
	\caption{Qualitative examples of point cloud registration by RANSAC with different descriptors on the 3DMatch, ModelNet40, and ETH datasets. The percentages denote the overlap ratio (OR) between fragments 1 and 2.}
	\label{fig:qualitative_results_3dmatch_modelnet_eth}
\end{figure*}

\begin{figure}[h]
	\centering
	\includegraphics[width=\linewidth]{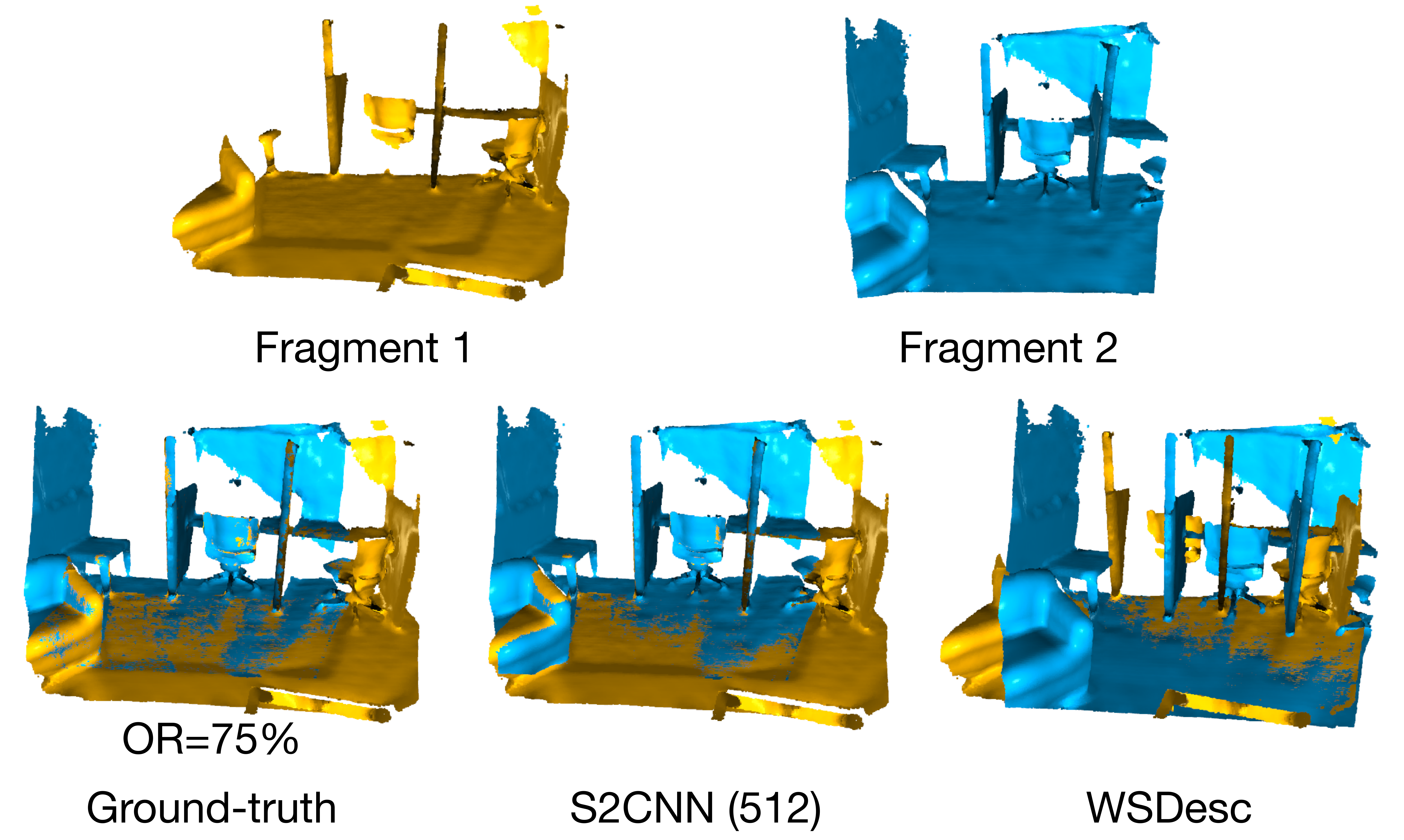}
	\caption{A failure case of point cloud registration with our local descriptor.}
	\label{fig:fail_cases_3dmatch}
\end{figure}

\textbf{Qualitative Visualization.}
\Figure{\ref{fig:qualitative_results_3dmatch_modelnet_eth}} (top) visualizes challenging point cloud registration examples with a large portion of flat surfaces, where our descriptor demonstrates better robustness.
\Figure{\ref{fig:fail_cases_3dmatch}} shows a failure case for our local descriptor, which we ascribe partly to the difficulty of extracting discriminative local descriptors for the thin structures of the two point clouds with the low-resolution ($\VoxelGridResolution = 16$) voxel grids.
More qualitative registration examples can be found in the supplementary material.

\textbf{Computation Time.}
\Table{\ref{tab:runtime_descriptors}} shows the running time comparisons.
The results were collected on a desktop computer with an Intel Core i7 @ 3.6GHz and an NVIDIA GTX 1080Ti GPU.
Note that for the input preparation, PPF-FoldNet and CapsuleNet compute point pair features; S2CNN performs spherical representation conversion; and \OurMethodName{} estimates LRFs.
Overall, our method has the best running time, while S2CNN is computationally much slower.

In \Table{\ref{tab:runtime_ransac_supp}}, we further collect the running time of RANSAC-based registration~\cite{Zhou:2018:Open3D} with different descriptors.
\emph{NN Query} denotes the search time for building a putative correspondence set (\Equation{\eqref{suppeq:PutativeCorrespondenceSet}}) with an efficient proximity search structure, such as a $k$-d tree.
S2CNN (32) and \OurMethodName{} have better efficiency due to the lower descriptor dimensionality.
\OurMethodName{} has the best RANSAC time owing to its high IR performance.

\begin{table}[h]
	\centering
	\caption{Average running time (ms) of local descriptor extraction on the 3DMatch dataset.}
	\resizebox{0.95\columnwidth}{!}{%
		\begin{tabular}{l|cccc}
			\toprule
			                                                & Dim.          & Input Prep.        & Inference           & Total        \\
			\midrule                                                                                                                      
			PPF-FoldNet                                     &         512   & 4.01               &  0.44               &  4.45        \\
			CapsuleNet                                      &         512   & 3.94               &  7.19               & 11.13        \\
			S2CNN                                           &         512   & 6.44               & 26.56               & 33.00        \\
			S2CNN                                           &         32    & 6.44               & 26.58               & 33.02        \\
			\OurMethodName                                  &         32    & 0.37               &  3.00               &  3.37        \\
			\bottomrule
		\end{tabular}
	}
	\label{tab:runtime_descriptors}
\end{table}

\begin{table}[h]
	\centering
	\caption{Average running time (ms) of pairwise registration on the 3DMatch dataset.}
	\resizebox{0.8\columnwidth}{!}{%
		\begin{tabular}{l|ccc}
			\toprule
			                                                & Dim.          & NN Query           & RANSAC              \\
			\midrule                                                                                                       
			PPF-FoldNet                                     &         512   &  9795.76           &  5.55               \\
			CapsuleNet                                      &         512   &  3652.68           &  6.78               \\
			S2CNN                                           &         512   & 13926.57           &  2.95               \\
			S2CNN                                           &         32    &   657.77           &  3.77               \\
			\OurMethodName                                  &         32    &   979.08           &  2.84               \\
			\bottomrule
		\end{tabular}
	}
	\label{tab:runtime_ransac_supp}
\end{table}

\subsection{Ablation Study}
\label{subsec:AblationStudy}
To get a better understanding of the contribution of the network components used by \OurMethodName{}, we perform an extensive ablation study on the 3DMatch dataset.

\begin{table}[h]
	\centering
	\caption{Ablation studies on 3DMatch. $\InlierRatioThresh = 0.2$ for FMR.}
	\resizebox{0.6\columnwidth}{!}{%
		\begin{tabular}{l|ccc}
			\toprule
			                                                                                    & IR$\uparrow$       & FMR$\uparrow$       & RR$\uparrow$        \\
			\midrule                                                                                                                                                                       
			\OurMethodName{}                                                                    &     \textbf{42.3}  &      \textbf{77.7}  &      \textbf{80.0}  \\
			\midrule
			w/o LLS                                                                             &             25.3   &              51.6   &              72.8   \\
			w/ LP                                                                               &             38.9   &              72.9   &              78.0   \\
			w/ GP                                                                               &             40.0   &              74.2   &              78.7   \\
			\midrule
			w/o LRF                                                                             &             21.6   &              39.5   &              50.8   \\
			\midrule
			w/o $\CorrespondenceWeightEntry_{i}$                                                &              6.4   &               5.4   &              31.2   \\
			w/o $\CorrespondenceWeightEntry_{i}^{f}$                                            &              8.0   &               6.4   &              45.6   \\
			w/o $\CorrespondenceWeightEntry_{i}^{sm}$                                           &             32.2   &              63.3   &              74.3   \\
			\midrule
			w/o $\LossOrthogonality$                                                            &             39.4   &              75.3   &              77.5   \\
			w/o $\LossCycle$                                                                    &             39.5   &              75.7   &              78.0   \\
			\bottomrule
		\end{tabular}
	}
	\label{tab:ablation}
\end{table}

\textbf{Learnable Local Support.}
First, to validate the effectiveness of using a learnable local support (\Section{\ref{subsec:DescriptorExtraction}}), we test a variant of \OurMethodName{} using the predefined fixed-size local support from 3DSmoothNet.
We keep the loss $\LossRegistration$ and other settings unchanged.
In \Table{\ref{tab:ablation}}, the performance of this variant is listed as \emph{w/o LLS}, which is worse than our full model.
This indicates that making the local support size optimizable enables the network to extract more informative descriptors.

Next, we discuss alternative designs for the learnable local support.
In \OurMethodName{}, we propose to treat the local support size $\VoxelGridScale$ as a universal learnable parameter.
One might consider employing a sub-network to estimate the support size.
We implemented two variants for such a design.
One variant is based on a local sub-network that estimates a local support size $\VoxelGridScale_{i}$ individually for each point $\PointP_{i} \in \PointCloudP$.
Specifically, to regress $\VoxelGridScale_{i}$, we feed the local patch $\Neighborhood_{i}$ of $\PointP_{i}$, cropped with the predefined radius $\NeighborhoodRadius$, to a mini-PointNet~\cite{Qi:2017:PointNetPlusPlus,Deng_2018_CVPR}.
Thus the local support size $\VoxelGridScale_{i}$ is dependent on the local geometry of $\PointP_{i}$, 
but the mini-PointNet may need to hallucinate the support size due to the missing of points outside $\Neighborhood_{i}$.
We list this variant as \emph{w/ LP} in \Table{\ref{tab:ablation}}.

The other variant is based on a global PointNet~\cite{Qi:2017:PointNetPlusPlus} for estimating a local support size $\VoxelGridScale_{i}$ individually for each $\PointP_{i}$.
Specifically, we use the semantic segmentation architecture of PointNet~\cite{Qi:2017:PointNetPlusPlus} for the estimation.
We feed the point cloud $\PointCloudP$ to the PointNet backbone, which regresses a scalar value $\VoxelGridScale_{i}$ for each point $\PointP_{i}$ in a single forward pass.
In this case, the local support size $\VoxelGridScale_{i}$ is coupled with the global scene structure (\Ie{}, $\PointCloudP$) and the local geometry of $\PointP_{i}$.
We list this variant as \emph{w/ GP} in \Table{\ref{tab:ablation}}.

It is observed that both the \emph{w/ LP} and \emph{w/ GP} variants perform worse than \OurMethodName{}, which uses a universal learnable local support size (converged to 1.004m after training).
Interestingly, we found that the sizes $\VoxelGridScale_{i}$ learned by \emph{w/ LP} are on average 1.719m ($\pm 5.0\times10^{-6}$), and the ones learned by \emph{w/ GP} are on average 1.701m ($\pm 2.1\times10^{-2}$).
The small variance phenomenon of the two variants further confirms our choice of using a universal learnable local support size.
This is possibly because robust descriptor matching typically requires optimized yet consistent support sizes for rigidly aligning point clouds, especially in the presence of significant partiality and noise.
Although the experimental results indicate that an exhaustive parameter search seems possible, we stress that it is computationally expensive and needs to be repeated on each used point cloud dataset.
Instead, our proposal offers higher efficiency for learning informative local descriptors in an end-to-end manner.

\textbf{Local Reference Frame.}
In \Section{\ref{subsec:DescriptorExtraction}}, we follow 3DSmoothNet to use the LRF estimation for ensuring rotation invariance in the descriptors.
We test a variant of \OurMethodName{} without the LRF estimation and report its performance as \emph{w/o LRF} in \Table{\ref{tab:ablation}}.
Unsurprisingly, without LRF, the local descriptors have reduced descriptiveness, leading to worse geometric registration performance.
Our finding echoes a similar observation of the LRF contribution in the 3DSmoothNet work.
During training, due to the lack of rotation invariance in the descriptors, the putative correspondences across point clouds can be spurious in the estimation of affine transformations between point clouds,
making $\LossRegistration$ ineffective to provide training signals.

\textbf{Transformation Estimation.}
To estimate the affine transformation, we adopt a weighted quadratic formulation in \Equation{\eqref{eq:LeastSquaresEnergy}}, where the weight $\CorrespondenceWeightEntry_{i}$ is comprised of two terms $\CorrespondenceWeightEntry_{i}^{f}$ and $\CorrespondenceWeightEntry_{i}^{sm}$ (\Equation{\eqref{eq:CorrespondenceWeight}}).
We test a variant of \OurMethodName{} without the weights (\Ie{}, by setting $\CorrespondenceWeightEntry_{i} = 1$), and we show its performance as \emph{w/o} $\CorrespondenceWeightEntry_{i}$ in \Table{\ref{tab:ablation}}.
Further, we remove one of the weight terms from $\CorrespondenceWeightEntry_{i}$ and list the performance of these two variants as \emph{w/o} $\CorrespondenceWeightEntry_{i}^{f}$ and \emph{w/o} $\CorrespondenceWeightEntry_{i}^{sm}$ in \Table{\ref{tab:ablation}}, respectively.
We observe that the network \emph{w/o} $\CorrespondenceWeightEntry_{i}$ fails on 3DMatch, and so does \emph{w/o} $\CorrespondenceWeightEntry_{i}^{f}$.
This is likely because without the weight terms, especially, $\CorrespondenceWeightEntry_{i}^{f}$ computed directly from the extracted descriptors, the gradients \Wrt{} $\LossRegistration$ can only pass through positions and the differentiable NN search, making it less effective to flow into the descriptor extractor for optimization~\cite{choy2020deep}.
As also shown by \emph{w/o} $\CorrespondenceWeightEntry_{i}^{sm}$, adding $\CorrespondenceWeightEntry_{i}^{f}$ to the network results in reasonable performance on 3DMatch.
Nevertheless, incorporating spectral matching indeed boosts the learning of descriptors and thus the registration performance.

\textbf{Registration Loss.}
Our training loss $\LossRegistration$ penalizes deviation from rigidity for the estimated affine transformations without requiring the ground-truth.
$\LossRegistration$ is comprised of the orthogonality loss $\LossOrthogonality$ and the cycle consistency loss $\LossCycle$.
We study the contribution of the two loss terms by removing one of them during training and keeping other settings unchanged.
In the bottom of \Table{\ref{tab:ablation}}, we show the results of the two experiments as \emph{w/o} $\LossOrthogonality$ and \emph{w/o} $\LossCycle$, respectively.
Note that there is noticeable performance degradation when using each loss term alone, and combining both loss terms produces the best results.

\subsection{ModelNet40 Dataset}
\label{subsec:ModelNet40Dataset}

We also perform comparisons with existing learning-based registration methods on the ModelNet40 benchmark introduced by~\cite{Wang:PRNet:2019}.
The dataset has 40 man-made object categories.
There are 9,843 point clouds for training and 2,468 point clouds for testing.
To generate point cloud pairs, a new point cloud is obtained by transforming each testing point cloud with a random rigid transformation.
The rotation angle along each axis is sampled in the range of $[0^{\circ}, 45^{\circ}]$, and the 3D translation offset is sampled in $[-0.5, 0.5]$.
To synthesize partial overlapping for a point cloud pair, 768 nearest neighbors of a randomly placed point in 3D space are collected in each point cloud.

\textbf{Metrics.}
Given rotations and translations estimated by a specific registration method, we follow~\cite{Wang:PRNet:2019} to compare them with the ground-truth by measuring root-mean-square error (RMSE) and coefficient of determination ($\text{R}^2$).
The rotation errors are computed with the Euler angle representation in degrees.

\textbf{Comparisons.}
In \Table{\ref{tab:modelnet40_align_err}} (bottom),
we evaluate existing axiomatic registration methods, including ICP~\cite{Besl:ICP:1992}, Go-ICP~\cite{yang2015go}, and FGR~\cite{zhou2016fast}, and the RANSAC-based registration methods (w/o GT) previously tested on 3DMatch, including PPF-FoldNet, CapsuleNet, S2CNN, and our \OurMethodName{}.
Besides, two recent unsupervised learning-based registration methods, ARL~\cite{Deng_2021_ICCV} and RMA-Net~\cite{Feng_2021_CVPR}, are considered\footnote{\url{https://github.com/Dengzhi-USTC/A-robust-registration-loss}}\footnote{\url{https://github.com/WanquanF/RMA-Net}}.
However, the released codebase for RMA-Net failed to produce reasonable results on the ModelNet40 benchmark, and thus we include only the comparison with ARL.
In \Table{\ref{tab:modelnet40_align_err}} (top), we also show the performance of learning-based registration methods \emph{with} supervision, including PointNetLK~\cite{yaoki2019pointnetlk}, DCP~\cite{Wang_2019_ICCV}, PRNet~\cite{Wang:PRNet:2019}, DeepGMR\cite{yuan2020deepgmr}, RPM-Net~\cite{Yew_2020_CVPR}, Predator~\cite{huang2020predator}, and OMNet~\cite{Xu_2021_ICCV}.
In \Table{\ref{tab:modelnet40_align_err}} (bottom), our \OurMethodName{} achieves the best performance across all of the computed metrics, compared to the other methods w/o GT, and is even on a par with the state-of-the-art RPM-Net, which is a supervised method highly specialized for object-centric datasets.
\Figure{\ref{fig:qualitative_results_3dmatch_modelnet_eth}} (middle) shows qualitative registration examples, where our descriptor leads to more accurate alignment.
We further perform comparisons on the noisy ModelNet40 benchmark by~\cite{Wang:PRNet:2019}.
\Table{\ref{tab:modelnet40_noise_align_err}} reports the registration results, and our \OurMethodName{} shows better robustness to noise.

\begin{table}[h]
\centering
\caption{Testing on unseen point clouds of the ModelNet40 dataset \Wrt{} rotation ($\mathbf{R}$) and translation ($\mathbf{t}$) estimations. The best and the second best results among the methods w/o GT are highlighted.}
\resizebox{\columnwidth}{!}{%
\begin{tabular}{l@{\hspace{0.1em}}|c|cccc}
\toprule
                                         &  GT       & RMSE($\mathbf{R}$)$\downarrow$ & R$^2$($\mathbf{R}$)$\uparrow$ & RMSE($\mathbf{t}$)$\downarrow$ & R$^2$($\mathbf{t}$)$\uparrow$ \\
\midrule                                                                                                                                                                                   
PointNetLK                               &  w/       &            16.735              &            -0.654             &             0.045              &              0.975            \\
DCP-v2                                   &  w/       &             6.709              &             0.732             &             0.027              &              0.991            \\
PRNet                                    &  w/       &             3.199              &             0.939             &             0.016              &              0.997            \\                    
DeepGMR                                  &  w/       &            19.156              &            -1.164             &             0.037              &              0.983            \\
RPM-Net                                  &  w/       &             1.290              &             0.990             &             0.005              &              1.000            \\
\rev{Predator}                           &  \rev{w/} &        \rev{1.875}             &        \rev{0.979}            &        \rev{0.017}             &        \rev{ 0.997}           \\
\rev{OMNet}                              &  \rev{w/} &        \rev{4.280}             &        \rev{0.891}            &        \rev{0.019}             &        \rev{ 0.996}           \\
\midrule                                                                                                                                                                                   
ICP                                      &  w/o      &            33.683              &            -5.696             &             0.293              &             -0.037            \\
Go-ICP                                   &  w/o      &            13.999              &            -0.157             &             0.033              &              0.987            \\
FGR                                      &  w/o      &            11.238              &             0.256             &             0.030              &              0.989            \\
\rev{ARL}                                &  \rev{w/o}&        \rev{8.527}             &        \rev{0.570}            &        \rev{0.029}             &        \rev{-0.046}           \\
PPF-FoldNet                              &  w/o      &             2.285              &             0.969             &  \underline{0.013}             &   \underline{0.998}           \\
CapsuleNet                               &  w/o      &  \underline{2.180}             &  \underline{0.972}            &  \underline{0.013}             &   \underline{0.998}           \\
S2CNN (512)                              &  w/o      &             3.069              &             0.944             &             0.017              &              0.997            \\
S2CNN (32)                               &  w/o      &             3.234              &             0.938             &             0.014              &   \underline{0.998}           \\
\OurMethodName                           &  w/o      &     \textbf{1.187}             &     \textbf{0.992}            &     \textbf{0.008}             &      \textbf{0.999}           \\
\bottomrule
\end{tabular}
}
\label{tab:modelnet40_align_err}
\end{table}

\begin{table}[h]
\centering
\caption{Testing on unseen point clouds of the ModelNet40 dataset augmented with Gaussian noise.}
\resizebox{\columnwidth}{!}{%
\begin{tabular}{l@{\hspace{0.1em}}|c|cccc}
\toprule
                                         & GT         & RMSE($\mathbf{R}$)$\downarrow$ & R$^2$($\mathbf{R}$)$\uparrow$ & RMSE($\mathbf{t}$)$\downarrow$ & R$^2$($\mathbf{t}$)$\uparrow$ \\
\midrule                                                                                                                                                                                    
PointNetLK                               & w/         &             19.939             &              -1.343           &             0.057              &              0.960            \\
DCP-v2                                   & w/         &              6.883             &               0.718           &             0.028              &              0.991            \\
PRNet                                    & w/         &              4.323             &               0.889           &             0.017              &              0.995            \\ 
DeepGMR                                  & w/         &             19.758             &              -1.299           &             0.030              &              0.989            \\
RPM-Net                                  & w/         &              1.870             &               0.979           &             0.011              &              0.998            \\
\rev{Predator}                           & \rev{w/}   &         \rev{1.893}            &          \rev{0.979}          &        \rev{0.009}             &         \rev{0.999}           \\
\rev{OMNet}                              & \rev{w/}   &         \rev{4.504}            &          \rev{0.880}          &        \rev{0.021}             &         \rev{0.995}           \\
\midrule                                                                                                                                                                                    
ICP                                      & w/o        &             35.067             &              -6.252           &             0.294              &             -0.045            \\
Go-ICP                                   & w/o        &             12.261             &               0.112           &             0.028              &   \underline{0.991}           \\
FGR                                      & w/o        &             27.653             &              -3.491           &             0.070              &              0.941            \\
\rev{ARL}                                & \rev{w/o}  &         \rev{7.973}            &          \rev{0.624}          &        \rev{0.027}             &         \rev{0.023}           \\
PPF-FoldNet                              & w/o        &   \underline{4.151}            &    \underline{0.899}          &             0.009              &      \textbf{0.999}           \\
CapsuleNet                               & w/o        &              4.274             &               0.893           &             0.009              &      \textbf{0.999}           \\
S2CNN (512)                              & w/o        &              5.221             &               0.840           &  \underline{0.007}             &      \textbf{0.999}           \\
S2CNN (32)                               & w/o        &              5.040             &               0.850           &             0.009              &      \textbf{0.999}           \\
\OurMethodName                           & w/o        &      \textbf{3.500}            &       \textbf{0.928}          &     \textbf{0.006}             &      \textbf{0.999}           \\
\bottomrule
\end{tabular}
}
\label{tab:modelnet40_noise_align_err}
\end{table}

\subsection{Generalization to ETH Dataset}
\label{subsec:ETHDataset}

To evaluate the generalization ability of our 3D local descriptors, we follow~\cite{Gojcic_2019_CVPR} to conduct experiments on the ETH dataset~\cite{Pomerleau:2012}.
This dataset consists of point clouds of four outdoor scenes, which are mostly laser scans of vegetation like trees and bushes.
The number of points per point cloud is $\sim$100K on average, and 5K keypoints are randomly sampled in each point cloud for matching.
For learning-based descriptors, to test their generalization ability, we directly reuse the networks trained on 3DMatch (\Section{\ref{subsec:3DMatchDataset}}) and use the ETH dataset only as a test set.

\textbf{Comparisons.}
\Table{\ref{tab:eth_recall}} shows comparisons for the 3D local descriptors (\Section{\ref{subsec:3DMatchDataset}}) in terms of the FMR metric ($\InlierRatioThresh = 0.05$).
We observe that \OurMethodName{} achieves better performance than the other methods w/o GT in \Table{\ref{tab:eth_recall}} (bottom), demonstrating the superior generalization ability.
\Figure{\ref{fig:qualitative_results_3dmatch_modelnet_eth}} (bottom) shows qualitative registration examples from this challenging outdoor-scene dataset.
We further provide the running time comparisons in \Table{\ref{tab:runtime_descriptors_eth}}, which were collected on a server with an Intel Xeon CPU @ 2.20GHz and an NVIDIA GeForce RTX 2080Ti GPU.
The speed of our method is comparable with PPF-FoldNet and influenced by the significantly increased number of points per point cloud, due to the formulation of differentiable voxelization in \Equation{\ref{eq:DifferentiableVoxelization}}.
We will investigate further optimizations on this network layer in future work.

\begin{table}[ht]
\centering
\caption{FMR performance (\%) on the ETH dataset ($\InlierRatioThresh = 0.05$).  The learned descriptors are only trained on the 3DMatch dataset.}
\resizebox{0.95\columnwidth}{!}{%
\begin{tabular}{l|c|cc|cc|c}
\toprule
                & GT         & \multicolumn{2}{c|}{Gazebo}          & \multicolumn{2}{c|}{Wood}            &                  \\
                &            & Sum.              & Wint.            & Sum.              & Aut.             & Avg.             \\
\midrule
3DMatch         & w/         &            22.8   &            8.7   &            22.4   &            13.9  &            16.9  \\
CGF             & w/         &            38.6   &            15.2  &            19.2   &            12.2  &            21.3  \\
3DSmoothNet     & w/         &            91.3   &            84.1  &            72.8   &            67.8  &            79.0  \\
FCGF            & w/         &            22.8   &            10.0  &            14.8   &            16.8  &            16.1  \\
D3Feat          & w/         &            85.9   &            63.0  &            49.6   &            48.0  &            61.6  \\
LMVD            & w/         &            85.3   &            72.0  &            84.0   &            78.3  &            79.9  \\
SpinNet         & w/         &            92.9   &            91.7  &            94.4   &            92.2  &            92.8  \\
\midrule
FPFH            & w/o        &            40.2   &            15.2  &            24.0   &            14.8  &            23.6  \\
SHOT            & w/o        &            73.9   &            45.7  & \underline{64.0}  & \underline{60.9} &            61.1  \\
PPF-FoldNet     & w/o        &            39.7   &            24.2  &            25.6   &            19.1  &            27.2  \\
CapsuleNet      & w/o        &            33.2   &            15.2  &            22.4   &            17.4  &            22.0  \\
S2CNN (512)     & w/o        &    \textbf{79.9}  & \underline{58.1} &            63.2   &            53.9  & \underline{63.8} \\
S2CNN (32)      & w/o        &            70.7   &            46.7  &            54.4   &            45.2  &            54.2  \\
\OurMethodName  & w/o        & \underline{78.3}  &    \textbf{77.2} &    \textbf{95.2}  &    \textbf{93.9} &    \textbf{86.1} \\
\bottomrule
\end{tabular}
}
\label{tab:eth_recall}
\end{table}

\begin{table}[h]
\centering
\caption{Average running time (ms) of local descriptor extraction on the ETH dataset.}
\resizebox{0.95\columnwidth}{!}{%
\begin{tabular}{l|cccc}
\toprule
                                                & Dim.          & Input Prep.        & Inference           & Total        \\
\midrule                                                                                                                    
PPF-FoldNet                                     &         512   &            16.65   &              0.30   &      16.95   \\
CapsuleNet                                      &         512   &            16.93   &              2.43   &      19.35   \\
S2CNN                                           &         512   &             5.61   &             17.56   &      23.17   \\
S2CNN                                           &         32    &             5.53   &             17.61   &      23.14   \\
\OurMethodName                                  &         32    &             9.50   &              6.75   &      16.25   \\
\bottomrule
\end{tabular}
}
\label{tab:runtime_descriptors_eth}
\end{table}

\section{Conclusion, Limitations \& Future Work}
\label{sec:Conclusion}

We have presented \OurMethodName{} to learn point descriptors for robust point cloud registration in a weakly supervised manner.
Our framework is built upon a voxel-based representation and 3D CNNs for descriptor extraction.
To enrich geometric information in the learned descriptors, we propose to learn the local support size in the online point-to-voxel conversion with differentiable voxelization.
We introduce a powerful weakly supervised registration loss to guide the learning of descriptors.
Extensive experiments show that our descriptors achieve superior performance on existing geometric registration benchmarks.

One limitation of our approach is that the differentiable voxelization layer might be a speed bottleneck for large-size point cloud inputs. 
This issue could be alleviated by applying down-sampling.
Besides, to handle thin-structure inputs, a higher resolution of voxel grids might be needed, which, though, will also increase the computation cost.

For future work, we will study the application of the differentiable voxelization to other 3D analysis tasks, such as object recognition~\cite{Qi_2016_CVPR}.
Besides, it would be interesting to combine our registration loss with other sub-tasks of point cloud registration, such as keypoint detection~\cite{Bai_2020_CVPR} or outlier filtering of correspondences~\cite{choy2020deep}.
It is also worth investigating the extension of our descriptor to the learning of unsupervised non-rigid shape matching~\cite{roufosse2019unsupervised}.

\ifCLASSOPTIONcompsoc
  \section*{Acknowledgments}
\else
  \section*{Acknowledgment}
\fi
The authors would like to thank the anonymous reviewers for their constructive comments.
Parts of this work were supported by the ERC Starting Grants No. 758800 (EXPROTEA), the ANR AI Chair AIGRETTE, and City University of Hong Kong (Project No. 7005729).

\appendix

\textbf{Spectral Matching.}
In \Section{3.2} - \emph{Transformation Estimation}, we use the spectral matching technique to compute the correspondence compatibility $\CorrespondenceWeightEntry_{i}^{sm}$ in \Equation{(4)}.
For completeness, we describe the algorithmic details in the following.

Given the putative correspondence set $\CorrespondenceSet$, spectral matching aims to maximize the following inter-cluster score
\begin{equation}
	\label{suppeq:SpectralMatchingObjective}
	\CorrespondenceWeight^{sm} = \ArgMax_{\CorrespondenceWeight^{sm}} \ (\CorrespondenceWeight^{sm})^{\top} \CompatibilityMatrix \CorrespondenceWeight^{sm},
\end{equation}
where $\CompatibilityMatrix \in \mathbb{R}^{|\CorrespondenceSet| \times |\CorrespondenceSet|}$ is a correspondence compatibility matrix, and $\CorrespondenceWeight^{sm} \in \mathbb{R}^{|\CorrespondenceSet|}$ is an indicator vector whose $i^{\text{th}}$ entry denotes the association of the correspondence $\Correspondence_{i} \in \CorrespondenceSet$ with the main inlier cluster.
The entry $\CompatibilityMatrix (\Correspondence_{i}, \Correspondence_{j})$ measures the \textit{consistency} between correspondences $\Correspondence_{i} = (\PointP_{i}, \PointQ_{i})$ and $\Correspondence_{j} = (\PointP_{j}, \PointQ_{j})$ in terms of length distortion.
For $i \neq j$, $\CompatibilityMatrix (\Correspondence_{i}, \Correspondence_{j})$ is defined as follows:
\begin{equation}
	\CompatibilityMatrix (\Correspondence_{i}, \Correspondence_{j}) = \big[ 1 - \frac{\CorrespondenceLengthDifference_{ij}^{2}}{ \sigma_{d}^{2}} \big]_{+}, \  \CorrespondenceLengthDifference_{ij} = \| \PointP_{i} - \PointP_{j} \|_2 - \| \PointQ_{i} - \PointQ_{j} \|_2,
\end{equation}
where $[\cdot]_{+} = \max(\cdot, 0)$, and $\sigma_{d} = 0.1$ controls the sensitivity to length distortion.
For $i = j$, $\CompatibilityMatrix (\Correspondence_{i}, \Correspondence_{j}) = 0$, because there is no information on an individual correspondence.
The entries of $\CompatibilityMatrix$ defined above are non-negative and increase as the length distortions between correspondences decrease.
Thus $\CompatibilityMatrix$ intuitively captures the compatibility between any two correspondences.
The principal eigenvector of $\CompatibilityMatrix$, under the constraint $\|\CorrespondenceWeight^{sm}\|_2 = 1$, maximizes the above inter-cluster score and can be computed efficiently by the power iteration algorithm as follows:
\begin{equation}
	\label{suppeq:PowerIteration}
	\CorrespondenceWeight^{sm}_{k + 1} = \dfrac{\CompatibilityMatrix \CorrespondenceWeight^{sm}_{k}}{\|\CompatibilityMatrix \CorrespondenceWeight^{sm}_{k}\|_2},
\end{equation}
where $k$ denotes the $k^{\text{th}}$ iteration.
We use $\CorrespondenceWeight^{sm}_{0} = \textbf{1}$.
In practice, we find that the power iteration algorithm converges in 10 iterations.
We assign the $i^{\text{th}}$ entry of $\CorrespondenceWeight^{sm}$ to $\CorrespondenceWeightEntry_{i}^{sm}$.

\begin{figure*}[h]
	\centering
	\includegraphics[width=\linewidth]{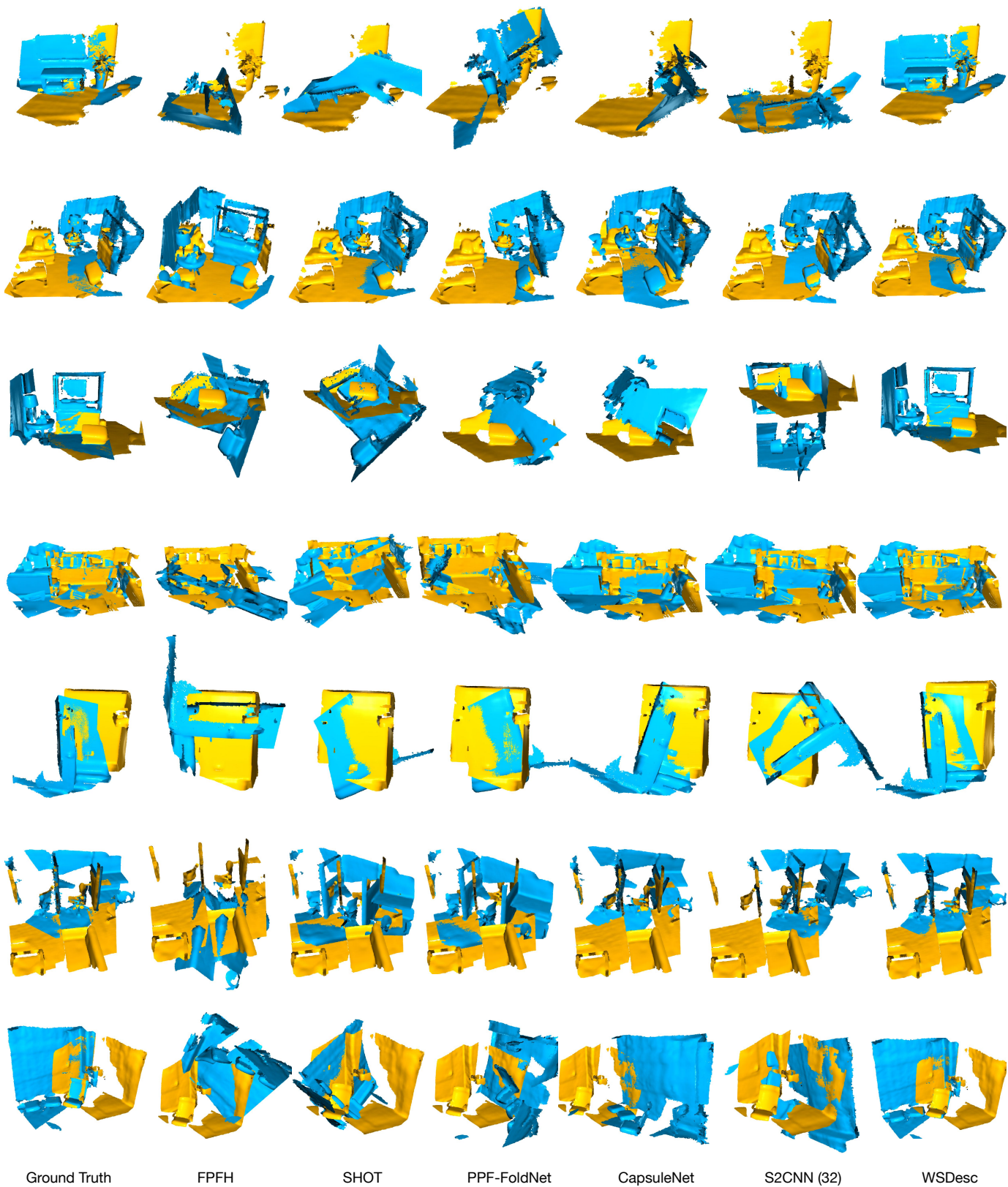}
	\caption{More point cloud registration results by RANSAC with different descriptors.}
	\label{fig:qualitative_results_3dmatch_supp}
\end{figure*}

\textbf{Qualitative Results.}
More qualitative registration results are shown in \Figure{\ref{fig:qualitative_results_3dmatch_supp}}.

\ifCLASSOPTIONcaptionsoff
  \newpage
\fi


\bibliographystyle{IEEEtran}
\bibliography{references}

\end{document}